%% file: scheduling_paper.tex
\documentclass{article}

\usepackage{microtype}
\usepackage{graphicx}
\usepackage{subcaption}

\usepackage{booktabs} 

\usepackage{hyperref}

\usepackage[accepted]{icml2025}

\usepackage{amsmath}
\usepackage{amssymb}
\usepackage{mathtools}
\usepackage{amsthm}
\usepackage{color}
\usepackage{xspace}
\usepackage{algorithm,algorithmic}
\usepackage{threeparttable}
\usepackage{breakurl}

\usepackage[capitalize,noabbrev]{cleveref}
\usepackage{enumitem,amssymb}
\newlist{todolist}{itemize}{2}
\setlist[todolist]{label=$\square$}
\theoremstyle{plain}

\theoremstyle{definition}

\theoremstyle{remark}

\usepackage[textsize=tiny]{todonotes}

\newif\ifcomm
\commtrue

\newcommand{\sys}{\texttt{SkipPipe}\xspace}
\newcommand{\cc}[1]{\texttt{CC#1}\xspace}
\newcommand{\tc}[1]{\texttt{TC#1}\xspace}
\newcommand{\hr}[1]{\textit{HR#1}\xspace}

\usepackage{soul}
\icmltitlerunning{\sys:  Partial Pipeline Parallelism}

\begin{document}

\twocolumn[
\icmltitle{\sys: Partial and Reordered Pipelining Framework for Training LLMs in Heterogeneous Networks}



\icmlsetsymbol{equal}{*}

\begin{icmlauthorlist}
\icmlauthor{Nikolay Blagoev}{wt}
\icmlauthor{Lydia Yiyu Chen}{un}
\icmlauthor{O\u{g}uzhan Ersoy}{ga}
\end{icmlauthorlist}

\icmlaffiliation{wt}{Worker Thread}
\icmlaffiliation{un}{University of Neuchatel \& TU Delft}
\icmlaffiliation{ga}{Gensyn}

\icmlcorrespondingauthor{Nikolay Blagoev}{workerthreadllc@gmail.com}

\icmlkeywords{LLM, Distributed Training, Scheduling}

\vskip 0.3in
]



\printAffiliationsAndNotice{}  

\begin{abstract}
 
Data and pipeline parallelism are ubiquitous for training of Large Language Models (LLM) on distributed nodes. Driven by the need for cost-effective training, recent work explores efficient communication arrangement for end to end training. Motivated by LLM's resistance to layer skipping and layer reordering, in this paper, we explore stage (several consecutive layers) skipping in pipeline training, and challenge the conventional practice of sequential pipeline execution. We derive convergence and throughput constraints (guidelines) for pipelining with skipping and swapping pipeline stages.
Based on these constraints, we propose \sys, the first partial pipeline framework to reduce the end-to-end training time for LLMs while preserving the convergence. The core of \sys is a path scheduling algorithm that optimizes the paths for individual microbatches and reduces 
idle time (due to microbatch collisions) on the distributed nodes, complying with the given stage skipping ratio.  We extensively evaluate \sys on LLaMa models from 500M to 8B parameters on up to 20 nodes. Our results show that \sys reduces training iteration time by up to 55\% compared to full pipeline.
 Our partial pipeline training also improves resistance to layer omission during inference, experiencing a drop in perplexity of only 7\% when running only half the model. Our code is available at \url{https://github.com/gensyn-ai/skippipe}.

\end{abstract}

\input{introduction}

\input{problem}

\input{method}

\input{evaluation}

\input{related}

\section{Conclusion}
Training state-of-the-art LLMs requires a significant number of GPUs and enormous training data.
There have been several work driven by the need for cost-effective training where they explored communication and computation improvements over DP and PP methods. 
Yet, existing PP methods are limited to the sequential execution of the layers.
In this paper we introduced a novel approach to pipeline parallelism, \sys, which makes use of stage skips and swaps to increase throughput by up to \(55\%\) in heterogeneous network settings. Our partial training also produces models resistant to layer removals during inference, which makes them suitable for early exit and fault tolerant inference. Our LLaMa-500M model trained with \sys experiences a drop in perplexity of only 7\% when running half the model.

Finally, while this paper focuses on the homogeneous nodes/ heterogeneous network, in future work, we plan to extend our solution to the full heterogeneous setting where nodes can have different memory and computational capacities.

\section*{Acknowledgment}
This work has been supported by Gensyn.

\bibliography{scheduling_paper}
\bibliographystyle{icml2025}

\newpage
\appendix
\onecolumn
\section{Model Configurations}
\label{appendix:mc}
\par We perform all our experiments with LLaMa-based \cite{llama} model architectures with the Sentence Piece Tokenizer \cite{SP}. The different models and their parameters are shown in Table \ref{table:models}. 

\begin{table}[h]
\caption{Model parameters.}
\label{table:models}
\begin{center}
\begin{tabular}[H]{p{6cm}p{1.2cm}p{1.2cm}p{1.2cm}p{1.2cm}}

  \toprule
  Model & Dim  & Heads & Layers & Context     \\ 
  
  \midrule
  LLaMa 50M & 288 & 6 & 12 & 256 \\
  LLaMa 500M & 1024 & 16 & 24 & 1024 \\
  LLaMa 1.5B~\cite{LayerSkip} & 2048  & 16 & 24 & 4096  \\
  LLaMa-2 7B~\cite{llama} & 4096  & 32 & 32 & 4096  \\
  LLaMa-3 8B~\cite{llama8b} & 4096  & 32 & 32 & 4096  \\
  
  \bottomrule

\end{tabular}
\end{center}
\end{table}

\section{Test Configurations}
\par Configurations of the throughput tests are presented in Table \ref{table:tests}. Stage sizes for the \(33\%\) case are (5,3,3,3,3,3) (6 stages, 5 of size 3 and 1 of size 5), and for the \(25\%\) case - (6,4,4,4) (4 stages, 3 of size 4, 1 of size 6).
\par For the convergence test, 4 samples per microbatch were used, with a total batch size of 737K tokens. Learning rate was set to \(3\times10^{-4}\) and gradient norms were clipped to \(1.0\).
\subsection{DT-FM-skip path selection}
\label{qrps}
\par Here we explain how the DT-FM-skip is determined. We choose paths that satisfy constraints \cc{1} in an optimised arrangement of nodes in stages. DT-FM-skip serves as a skip baseline which is mainly optimised for the initial node arrangement, but not necessarily for the partial microbatch paths.
\par In order to keep comparison fair, we chose to satisfy constraints \tc{1}, as otherwise delays will be introduced on nodes whose memory is exceeded, as it will need to wait for a backwards pass to come through, before it can continue with this forward pass. Due to this, and our experiment setups, we also inadvertently would satisfy constraints \cc{3}. Thus the algorithm for determining the paths for this baseline is identical to that of the non-collision aware one, except that the computation time of each node and communication time between nodes is set to \(1\). Thus the algorithm does not optimise for fastest paths or \tc{2} constraints.
\begin{table}
\caption{Test settings.}
\label{table:tests}
\begin{center}
\begin{threeparttable}[b]
\begin{tabular}[H]{p{1cm}p{4.2cm}p{1.4cm}p{1.2cm}p{1.6cm}}
  \toprule
  Skip & Path finding  & World size & Samples per MB & Batch size  (tokens)   \\ 
  \midrule
  0\% & DTFM \cite{dtfm}  & $18\rightarrow 20$\tnote{a} & 1 & 184K  \\
  \hline
  0\% & DTFM \cite{dtfm}  & $18\rightarrow 20$\tnote{a} & 2 & 368K  \\
  \hline
  0\% & DTFM \cite{dtfm}  & $18\rightarrow 20$\tnote{a} & 4 & 737K  \\
  \hline
  33\% & DT-FM-skip   & 20 & 1 & 184K  \\
  \hline
  
  33\% & DT-FM-skip  & 20 & 2 & 368K  \\
  \hline
  33\% & DT-FM-skip   & 20 & 4 & 737K  \\
  \hline
  33\% & non-collision aware  & 20 & 1 & 184K  \\
  \hline
  
  33\% & non-collision aware  & 20 & 2 & 368K  \\
  \hline
  33\% & non-collision aware  & 20 & 4 & 737K  \\
  \hline
  33\% & collision aware  & 20 & 1 & 184K  \\
  \hline
  
  33\% & collision aware  & 20 & 2 & 368K  \\
  \hline
  33\% & collision aware  & 20 & 4 & 737K  \\
  \hline
  0\% & DTFM \cite{dtfm}  & $16\rightarrow 18$\tnote{b} & 1 & 147K  \\
  \hline
  0\% & DTFM \cite{dtfm}  & $16\rightarrow 18$\tnote{b} & 2 & 294K  \\
  \hline
  0\% & DTFM \cite{dtfm}  & $16\rightarrow 18$\tnote{b} & 4 & 589K  \\
  \hline
  25\% & DT-FM-skip  & 18 & 1 & 147K  \\
  \hline
  
  25\% & DT-FM-skip   & 18 & 2 & 294K  \\
  \hline
  25\% & DT-FM-skip  & 18 & 4 & 589K  \\
  \hline
  25\% & non-collision aware  & 18 & 1 & 147K  \\
  \hline
  
  25\% & non-collision aware  & 18 & 2 & 294K  \\
  \hline
  25\% & non-collision aware  & 18 & 4 & 589K  \\
  \hline
  25\% & collision aware  & 18 & 1 & 147K  \\
  \hline
  
  25\% & collision aware  & 18 & 2 & 294K  \\
  \hline
  25\% & collision aware  & 18 & 4 & 589K  \\
  \bottomrule
\end{tabular}
\begin{tablenotes}
\item [a] In 33\% skip experiment, we use 6 stages with $(5,3,3,3,3,3)$ nodes. DT-FM 0\% skip does not use extra nodes in the first stage (as all stages are used equally). To (over)compensate them using less nodes (while keeping the stage sizes the same), we project their performance by linearly reducing the latency accordingly. In other words, if an iteration of DT-FM case takes 20sn with 18 nodes, we assume it would take 18sn with 20 nodes. Considering the communication of those additional nodes being ignored, this is upper bound of their performance.
\item [b] Same with above except in 25\% skip experiment, we use 4 stages with $(6,4,4,4)$ nodes. Therefore, 16 nodes are projected to 18 nodes.
\end{tablenotes}
\end{threeparttable}
\end{center}
\end{table}

\section{Detailed Path Selection Algorithm}\label{sec:appx_algorithm}
In Algorithm~\ref{alg:csf} we present the steps of our path selection function.

\begin{algorithm}
\caption{Path Selection Function.}
\label{alg:csf}
\begin{algorithmic}[1]
{\small
\REQUIRE{ \(\mathcal{S}\), \(k\%\), $G$ - initial node/stage arrangement}
\ENSURE{\(\mathcal{P}\)} 
\STATE \(\mathcal{O} \leftarrow \emptyset\)
\STATE $ T_{constraints} \leftarrow \emptyset$
\STATE Assign $S_0$ to the first stage of $ T_{paths}$
\STATE $ T_{paths} \leftarrow$ find paths via A*(\(G, T_{constraints}\)) 
\STATE Order $ T_{paths}$ by their time to complete in ascending order
\STATE \(T_{cost} \leftarrow \) time for slowest agent to complete route
\STATE Insert \(T\) into \(Open\)

\WHILE{$|\mathcal{O}|<32$ }
\STATE \(T \leftarrow \) best solution from \(Open\)
\STATE Check for \(S_i\) in \(T\) which has more than \(|\mathcal{S}|k\%\) agents going through other than \(S_0\)

\IF{conflict}
\STATE \(\mathcal{K} \leftarrow\) number of agents going through \(S_i\)
\STATE \(Solution \leftarrow\) new node
\STATE $ Solution_{constraints} \leftarrow T_{constraints}$
\FOR{each \(\mathcal{D}_m \in S_i\)}
\FOR{each of the \(\mathcal{K} - |\mathcal{P}|k\%\) fastest paths \(p \in \mathcal{P}\) going through \(S_i\)} 
\STATE $ Solution_{constraints} \leftarrow Solution_{constraints} + (p, -\inf, \inf, \mathcal{D}_m)$ 
\ENDFOR
\ENDFOR
\STATE $ Solution_{paths} \leftarrow$ find paths via A*(\(G, Solution_{constraints}\))
\STATE Order $ Solution_{paths}$ by their time to complete in ascending order
\STATE \(Solution_{cost} \leftarrow \) time for slowest agent to complete route
\STATE Insert \(Solution\) into \(Open\)
\ELSE
\STATE \(\mathcal{O} \leftarrow \mathcal{O} \cup T\)
\ENDIF
\ENDWHILE
\WHILE{\(\mathcal{O}\) is not empty}
\STATE \(T \leftarrow \) best solution from \(\mathcal{O}\)
\STATE Check for conflicts \tc{1} or \tc{2} in \(T\)
\IF{conflict of type \tc{1}}
\STATE \(D_k\) would be the node, whose \(m\) is exceeded as per \tc{1}
\STATE \(\mathcal{K}\) the paths that go through \(D_m\)
\STATE \(Solution \leftarrow\) new node
\FOR{each of the \(\mathcal{K} - m\) fastest paths \(p \in \mathcal{P}\) going through \(D_k\)} 
\STATE $ Solution_{constraints} \leftarrow Solution_{constraints} + (p, -\inf, \inf, \mathcal{D}_k)$ 
\ENDFOR

\STATE $ Solution_{paths} \leftarrow$ find paths via A*(\(G, Solution_{constraints}\))
\STATE Order $ Solution_{paths}$ by their time to complete in ascending order
\STATE \(Solution_{cost} \leftarrow \) time for slowest agent to complete route
\STATE Insert \(Solution\) into \(\mathcal{O}\)
\ELSIF{conflict of type \tc{2}}

\STATE Two paths, \(p_i\) and \(p_j\) collide on \(D_k\). Each of them is at the node during the intervals \(t_{s,i},t_{e,i}\) and \(t_{s,j},t_{e,j}\), respectively
\STATE \(Solution \leftarrow\) new node
\IF{\( E2E(p_i) >  E2E(p_j)\) or \(|E2E(p_j) - E2E(p_j)| < \delta \)}
\STATE \(Solution_{constraints} \leftarrow T_{constraints} + (p_j,t_{s,i},t_{e,i}, D_k)\)
\ENDIF
\IF{\( E2E(p_i) <  E2E(p_j)\) or \(|E2E(p_j) - E2E(p_j)| < \delta \)}
\STATE \(Solution_{constraints} \leftarrow T_{constraints} + (p_i,t_{s,j},t_{e,j}, D_k)\)
\ENDIF
\STATE $ Solution_{paths} \leftarrow$ find paths via A*(\(G, Solution_{constraints}\))
\STATE Order $ Solution_{paths}$ by their time to complete in ascending order
\STATE \(Solution_{cost} \leftarrow \) time for slowest agent to complete route
\STATE Insert \(Solution\) into \(\mathcal{O}\)
\ELSE
\STATE \textbf{Return} $\mathcal{P} \leftarrow T_{paths}$
\ENDIF
\ENDWHILE
\STATE \textbf{Return} $\emptyset$
}
\end{algorithmic}
\end{algorithm}

\section{Possible Extensions of Our Algorithm}
\subsection{Path Coarsening}
\par Here we also present an alternative path finding method based on path coarsening that finds solutions faster, but they may be sub-optimal. The reason for the sub-optimality is that it may increase idle time on devices. However, in a strictly homogeneous device memory setting, it can ignore \tc{2} constraints. Thus, it is best suited for large systems of nodes with equal memory capabilities, where an exact solution may be too costly to compute and due to the homogeneity of the system, most quality solutions will have similar throughput.

\par Here we make use of \textbf{path coarsening} - grouping multiple paths into one meta-agent. Meta-agents traverse a node sequentially, without interruption, and take the total amount of execution time of all the microbatches in the meta-agent. Meta-agent thus become 2-dimensional objects, rather than the point-agents we were considering prior. The downside is that in heterogeneous environments, meta-agents might become more stretched out or mode condensed as they traverse the system. Consider three nodes arranged as A-B-C, taking time to process a microbatch of respectively 1, 2, and 1 seconds. communication between them is 1 second per microbatch. Initially, a meta-agent of 2 microbatches, would have a size of 2 seconds at node A. At node B, due to its delay of processing, the agent will be resized to size of 4, even though the subsequent node would have a gap of 1 second where it would be idle between the two microbatches. However, with meta-agents with multiple paths this level of detail is lost in favour of faster solutions. The best benefit of path coarsening is in a fully homogeneous node setting - equal processing time and equal memory for each. In such a setting we can create meta-agents with number of microbatches in them equal to the memory of the nodes. When finding the solution, all meta-agents will have mutually exclusive paths, thus no collisions need to be considered. Proving the optimality of such a solution is beyond the scope of the paper.
\par In fact our solution has already made use of a degree of coarsening, as we optimise only the first forward pass in an iteration. It is possible to find an even better solution across where no path is reused by microbatches, however, due to the difficulty of finding such a solution even for a small world and small number of agents, we have not performed further analysis.

\subsection{Multiple Swaps}
It is possible to increase the number of swaps by introducing some linear penalty for paths that have swaps more than the desired amount, as a higher number of swaps  hampers convergence, but may increase throughput. It is also possible to define an additional constraint that sets a maximum number of swaps across all paths, which would be delegated to CBS to resolve like constraint \cc{3}, e.g. at most \(|\mathcal{P}|\) swaps across all paths. This would however greatly increase the time to find a quality solution. 
\section{Further Experimental Results}
\par
\begin{figure}[htp]
	\centering
	\begin{tabular}{cc}
		\includegraphics[width=0.485\columnwidth]{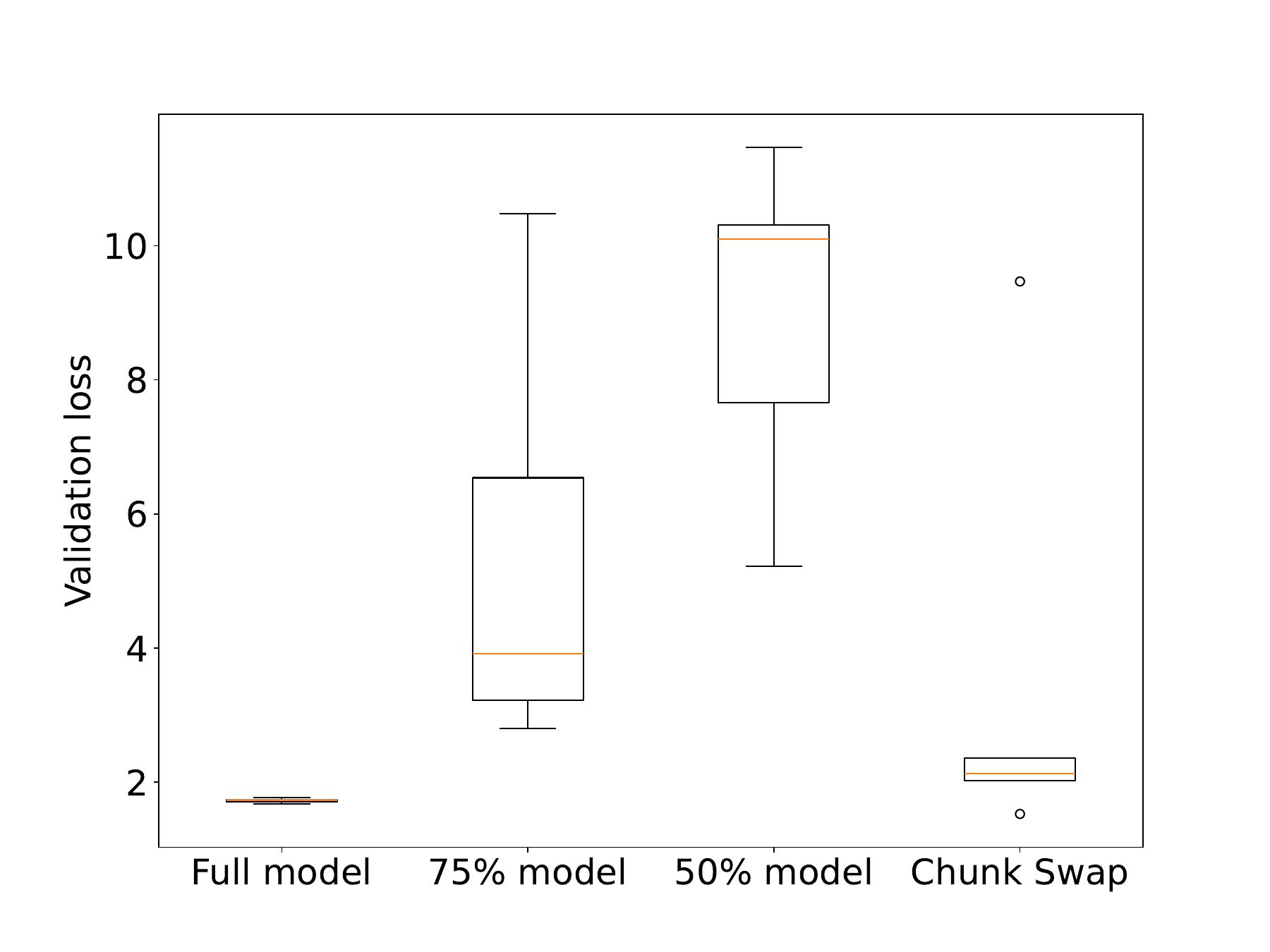} &
		\includegraphics[width=0.485\columnwidth]{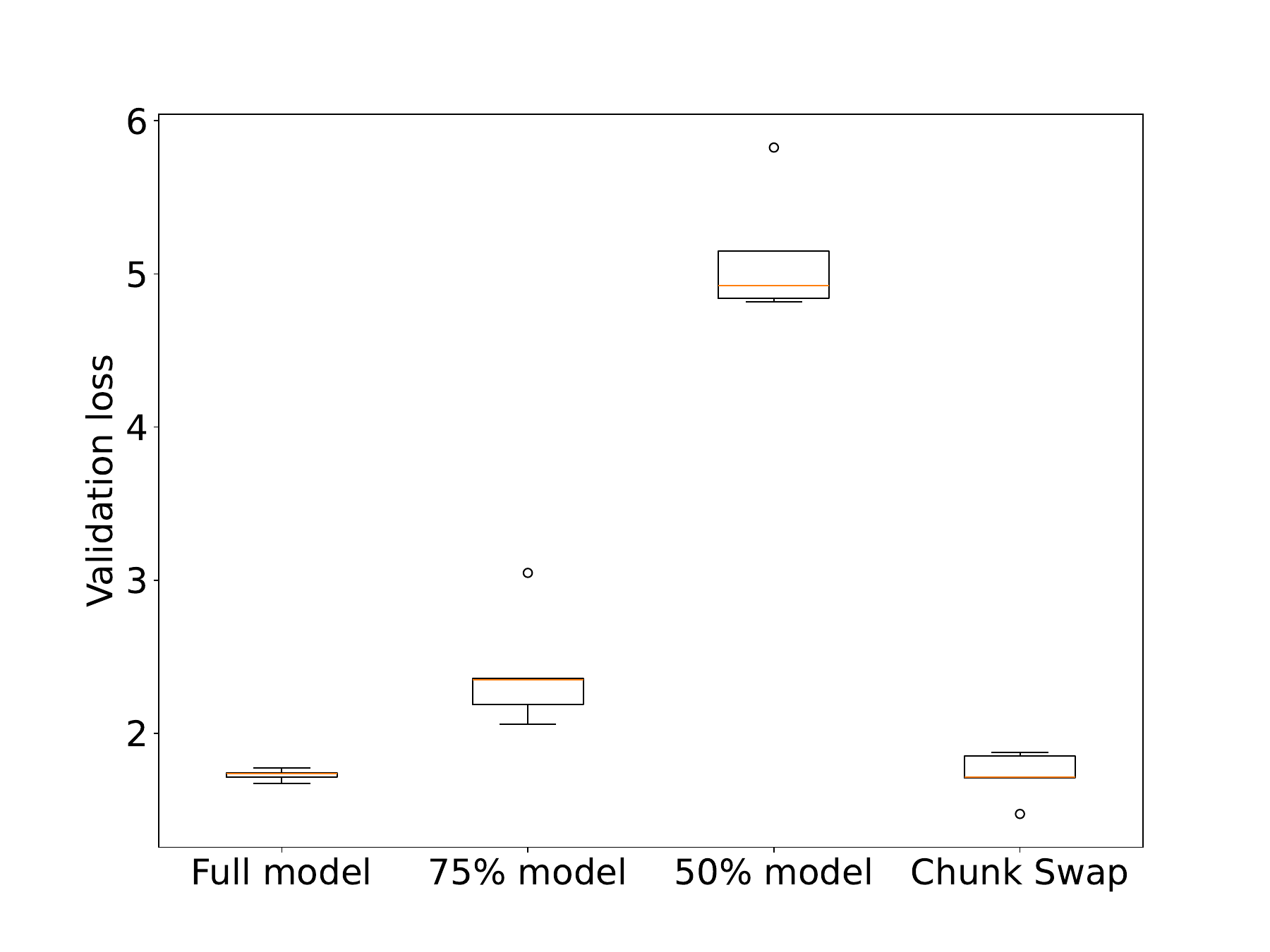} \\
		
		(a) Fully random skipping. & (b) Fixed first four layers.  
	\end{tabular}

		\caption{The validation loss of LLaMa-7B under \% of random skipping in pipeline training.}
		\label{fig:inferencellmfixed4}
	\end{figure}

\begin{figure}
    \centering
    \includegraphics[width=\linewidth]{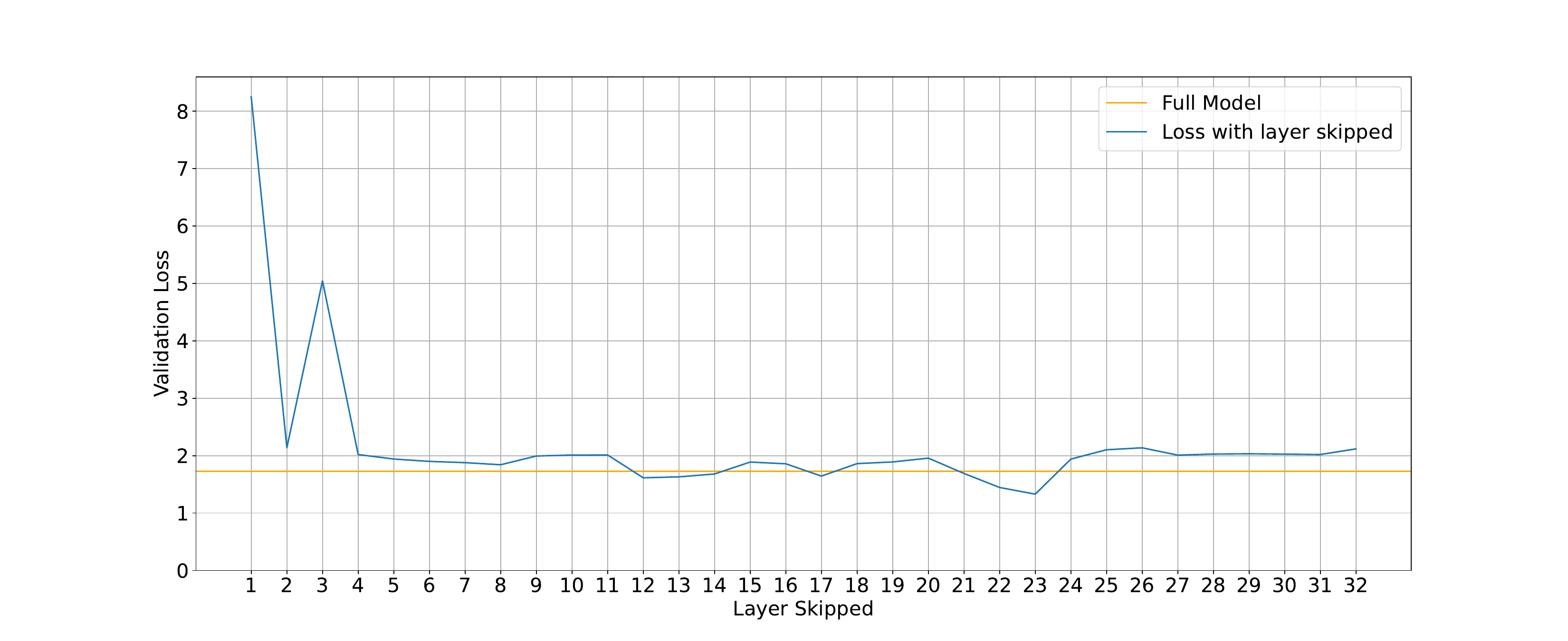}
    \caption{Validation loss when a given layer is skipped.}
    \label{fig:layers-skip}
\end{figure}

 Here we reaffirm our findings from Section \ref{sec:31} in the context of Large Language Models used in practice. While training billion parameter models is too expensive, here we focus on the inference case to confirm some of our previous findings. To such an end, we conduct an empirical performance study on skipping layers during inference on training a LLaMa-7B model~\cite{llama}, on the WikiPedia dataset \cite{wikidump}. We consider four layer skipping strategies: (i) 0\% skipping running the entire model end to end, (ii) 25\% random skipping, (iii) 50\% of random skipping, and (iv) 0\% skipping and swapping the order of two chunks of size 4. We also repeat these four strategies by fixing the first four layer (they never get skipped or swapped). We summarize their loss  in figure Fig. \ref{fig:inferencellmfixed4}. Additionally, we demonstrate in the same setting the effect on inference of skipping any arbitrary stage in the LLaMa-7B model \cite{llama} during inference in \ref{fig:layers-skip}.

\end{document}

%% file: introduction.tex
\section{Introduction}
Deep transformer-based architectures~\cite{transformer} have recently enabled unprecedented performance on complex language and cognitive tasks~\cite{gpt}. 
These leaps can be explained by the ever growing corpora of available data and by the increasing size of (Large) Language Models (LLMs)~\cite{llama,gpt3,palm,BERT,megatron}. As a consequence, models are now too large to fit and be efficiently trained on a single GPU.

Distributed training techniques, such as Pipeline Parallelism (PP) and Data Parallelism (DP), become indispensable to efficiently  train large models on distributed nodes.\footnote{Here, nodes are also referred as devices or GPUs.} In the former the model is split in stages, containing non-overlapping sections of the model, across a set of nodes, which communicate sequentially between each other to run the whole model, thus forming a pipeline. In the latter, multiple pipelines train the model independently on different data batches, communicating between each other to synchronize the model weights after an update. Training with the standard synchronous algorithms and renting private clusters to train models can easily cost more than tens of thousands of dollars~\cite{dtfm}, even for smaller models. Some prior work has proposed training on smaller clusters over a heterogeneous network (different communication latency and bandwidth between nodes), however in such a setting the communication between the GPUs is still  one of the main limiting factors~\cite{dtfm}.

Recent work has aimed to improve cost effectiveness of LLM training via less frequent synchronizations within DP~\cite{diloco,demo,opendiloco} and heterogeneity-aware arrangement of the GPUs/clusters ~\cite{dtfm,swarm,hetpipe,metis,flashflex}.
The former, Diloco~\cite{diloco} and DeMo~\cite{demo}, show that the synchronization for gradients can be reduced by orders of magnitude. The latter, heterogeneity-aware scheduling methods~\cite{dtfm,swarm,hetpipe,metis,flashflex}, present efficient arrangement of the GPUs to minimize the communication overhead in heterogeneous network settings. 
Yet, pipelining is done strictly following a sequential execution of layers from beginning to the end for all microbatches~\cite{gpipe,zb,dtfm,hetpipe}.

 \begin{figure*}[t]
    \centering
\includegraphics[width=2\columnwidth]{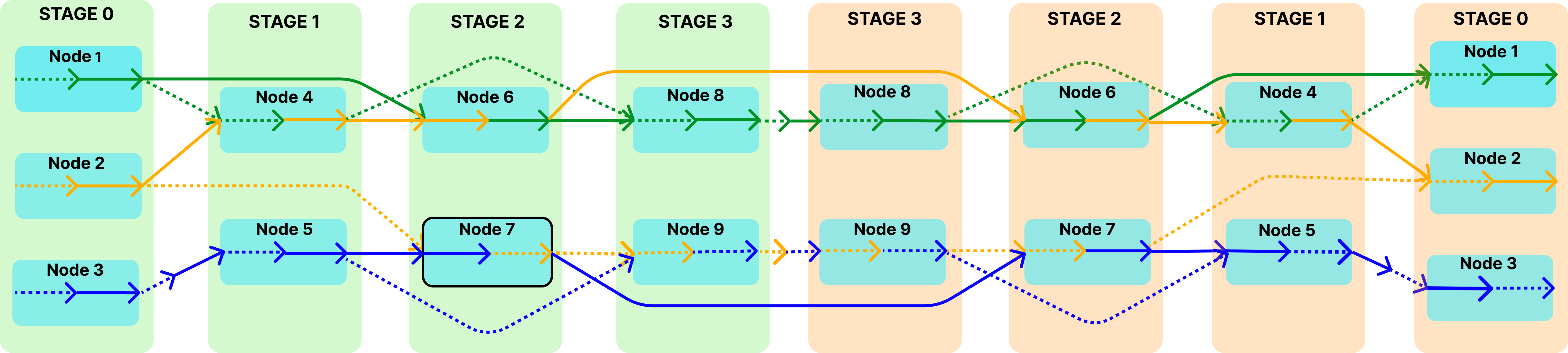}
      \caption{An example of   
      partial pipeline parallelism scheduling where each colored (solid or dashed) path represents a different microbatch. Each node in stage 0 sends out 2 microbatches, the first in solid, the second in dashed. Green backgrounds show the forward pass, while light orange - the backwards pass. For better visualisation, we omit the loss and deembedding computations. We place arrows to show the prioritisation of the microbatches from forward to backward pass within the same node. An example of a collision can be seen on node 7 during the forward pass, which subsequently delays the execution of the solid blue microbatch because of the dashed yellow microbatch.}   \label{fig:highlevelourprotocol}
\end{figure*}

The works of \cite{vitrob,LayerDrop,LayerSkip} have demonstrated transformer architectures' robustness against layer skipping and even layer reordering during training and inference. We leverage this fact to propose a novel optimisation to traditional training - \sys, which is the first partial pipeline framework that skips (and re-orders) pipeline stages. \sys improves the training throughput measured in time per iteration while preserving convergence of the model on distributed nodes and it is also suitable for communication heterogeneous settings. Moreover, the partial training via stage skipping in \sys also improves the inference with layer/stage skips, which is beneficial for not only the early-exit inference methods, but also fault-tolerant ones.

To minimize the end-to-end training latency via stage skipping and reordering, \sys is composed of two modules: allocating nodes to pipeline stages, and a path scheduler for microbatches.
For a given (heterogeneous) network of nodes and pipeline stages of an LLM model, \sys first allocates nodes to the stages where nodes in the same stage communicate in DP manner and nodes in different stages communicate in PP. Then, differently from standard pipelining where each microbatch passes through all stages sequentially along the same path, \sys 
schedules partial paths for each microbatch that skip some stages and run others out of order. 
As illustrated in Figure~\ref{fig:highlevelourprotocol}, each microbatch skips \(k\%\) of the model where \(k\) is a user-defined parameter.

The key challenge is how to select the path such that the number of microbatch collisions is minimised and the model convergence is not affected negatively.

Our contributions can be summarised as follows: 
\begin{itemize}
    \item We propose a novel and effective partial and reordered pipelining framework for distributed LLM training to reduce the computation and communication overhead.
    \item We design a pipeline execution scheduler optimising the throughput for heterogenous network of nodes by utilising skipping and swapping stages and reducing collisions (overlapping microbatches executions). 
    \item We evaluate our scheduler and present up to \(60\%\) reduction in training time when training with \sys compared to training with a standard full-model framework in a heterogeneous network. Also, we show that there is no convergence degradation.
    \item We show that the models trained with \sys also provide significant resistance to layer omission during inference, e.g., with a perplexity drop off of only 7\% when executing half the model.
\end{itemize}

%% file: problem.tex
\section{System Setting}
In this section, we explain the distributed training environment and the terminology used throughout the paper.

\textbf{System setup.} There are $\mathcal{N}$ distributed nodes for training an LLM model of \(\mathcal{L}\) layers, which is divided in pipeline stages $\mathcal{S}:=(S_0, S_1,\ldots,S_s)$. 
Each stage \(S_i\) holds an (equal)\footnote{Not necessary for our solution, but for simplicity and clarity we focus on the homogeneous stage/node setting.} number of consecutive  layers \(L_j...L_{j+\delta}\) and there are no overlapping layers across stages.

We assume each node has the same memory capacity that allows them to operate the same number of microbatches. We assume each node can communicate with any other and the communication cost between nodes is modelled with $(\mathcal{B},\Lambda)$ matrices where communication between nodes \(N_i\) and \(N_j\) has a cost associated to it modelled by the latency \(\lambda_{i,j} \in \Lambda\) and bandwidth \(\beta_{i,j} \in \mathcal{B}\). Thus for a message of size $|msg|$, its communication takes \(\lambda_{i,j} + \frac{|msg|}{\beta_{i,j}}\) (mili)seconds. While communication may not be symmetric, since each link is used twice, once during forward and once during backward, we model latencies and bandwidth as the average of the two directions (e.g., \(\lambda'_{i,j} = \frac{\lambda_{i,j} + \lambda_{j,i}}{2}\)), as in \cite{dtfm}.

\textbf{Distributed Training.} Each node is mapped to a single LLM stage. 
To train the LLM with data and pipeline parallelism, a batch is split into multiple microbatches, which perform forward and backward passes through each stage. 
Nodes sharing the same stage communicate the gradient updates in DP, and nodes in different stages communicate activations in PP.

We consider synchronous updates in pipelining, i.e., the weight update of an iteration is done after all the corresponding microbatches are processed. 
However, unlike common pipelining where each microbatch passes through all stages in the sequential order, we propose partial and reordered pipelining which is explained below.

\textbf{Partial and Reordered Pipeline.} 
The prior work pinpointed that transformer-based architectures are robust to layer skipping, i.e., not executing a given layer ~\cite{vitrob,LayerDrop}. We investigate if stage skipping is also advantageous in pipeline parallelism. We term such an idea partial pipeline parallelism. In the full pipeline scenario, microbatches traverse through the stages sequentially, e.g. $\mathcal{S}:=(S_0, S_1,S_2,S_3,S_4,S_5)$. In our case microbatches can traverse through different sequence of stages, due to skipping a given stage ($\mathcal{S}:=(S_0,S_1,S_4,S_6)$) or swapping the order of two stages ($\mathcal{S}:=(S_0,S_1,S_3,S_2)$). The key research questions thus are which stages to skip such that negligible performance loss occurs to the LLM, while minimising the training time, by choosing faster and shorter paths through the system.

%% file: method.tex
\section{\sys}

In this section we present a novel approach to pipeline parallelism, employing skipping and swapping to reduce the required resources and increase throughput without degrading the training performance for LLMs. 
The goal is to find a viable partial pipeline schedule (paths of the microbatches) that minimizes the overall training latency given the throughput target.
Before going into our scheduler, first we derive the convergence and throughput guidelines for partial pipelining.

\textbf{Partial pipeline schedule.} Given a DP and PP arrangement of nodes (constituting a graph) with the given communication and computation limitations per link and node respectively, we need to find paths  \(p_1, p_2... \in \mathcal{P}\) (of a sequence of nodes) for each microbatches such that the time to complete a forward and a backward pass through them is minimized, i.e., end-to-end latency for training a batch of data. 

Each path $p_i$ travels a sequence of nodes  from the starting node back to itself (considering forward and backward), such that only \(k\%\) of stages are skipped (and no stages are repeated in the path). The ordering of nodes in the backward pass needs to be the same as in the forward one.
A path $p_i$ can be represented with respect to the stages ($p_i:= S_{i_1},\ldots,S_{i_l}$) or the nodes ($p_i:= N_{i_1},\ldots,N_{i_l} $) that it passes through where $l:=(100-k)\%$ of the stages. 

\subsection{Guideline for Partial Pipelining Scheduler}
\label{sec:31}

Here, we explain our guideline for a partial pipeline scheduler that selects the paths for each microbatch through a motivation example. 
We present three convergence and two throughput constraints to optimize the path selection. We derive the convergence constraints from our experimental results and previous work and the throughput constraints are based on the node and network limitations.

\begin{figure*}[t]
\begin{tabular}{ccc}
	\includegraphics[width=\columnwidth]{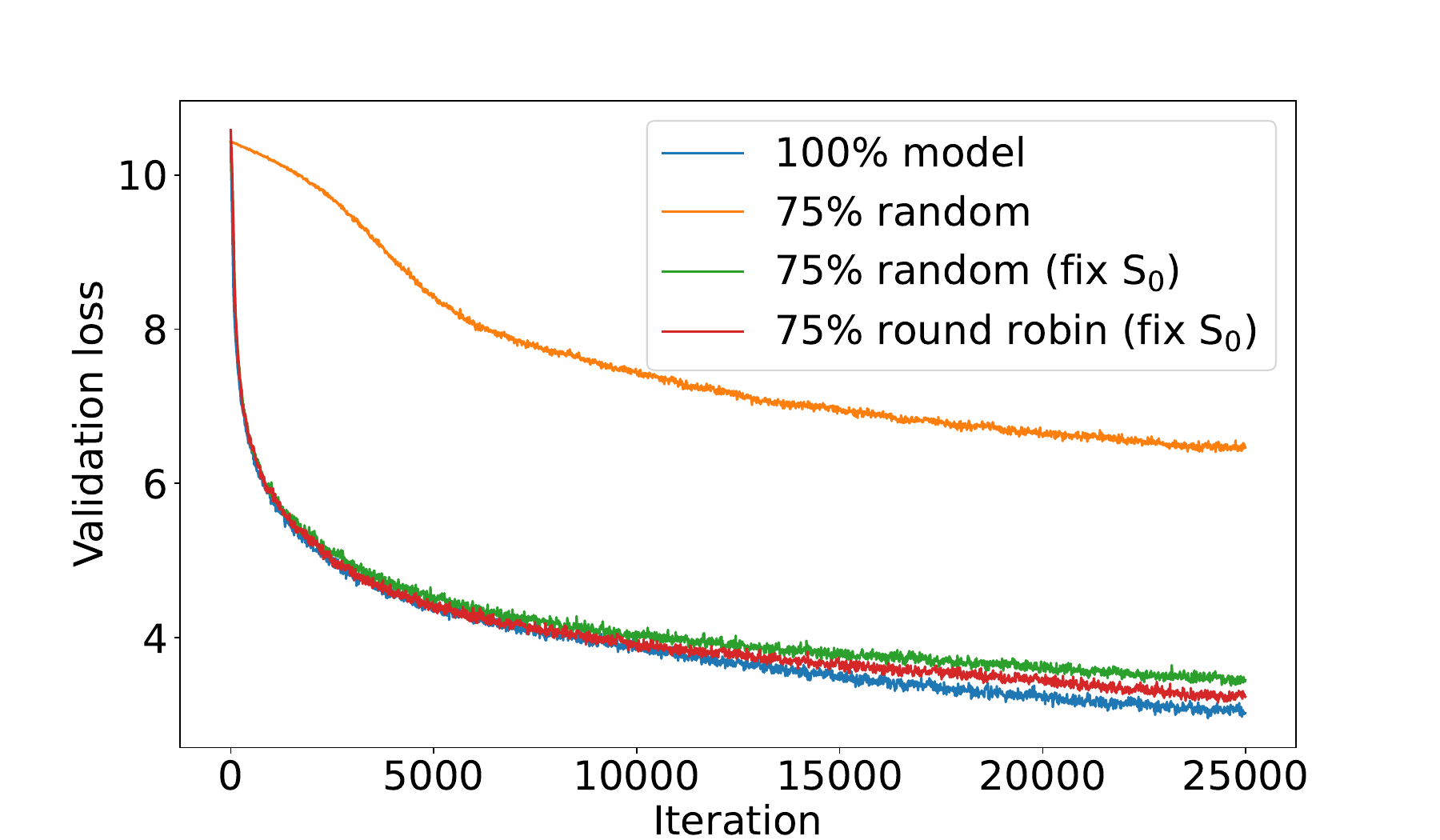} &
\includegraphics[width=\columnwidth]{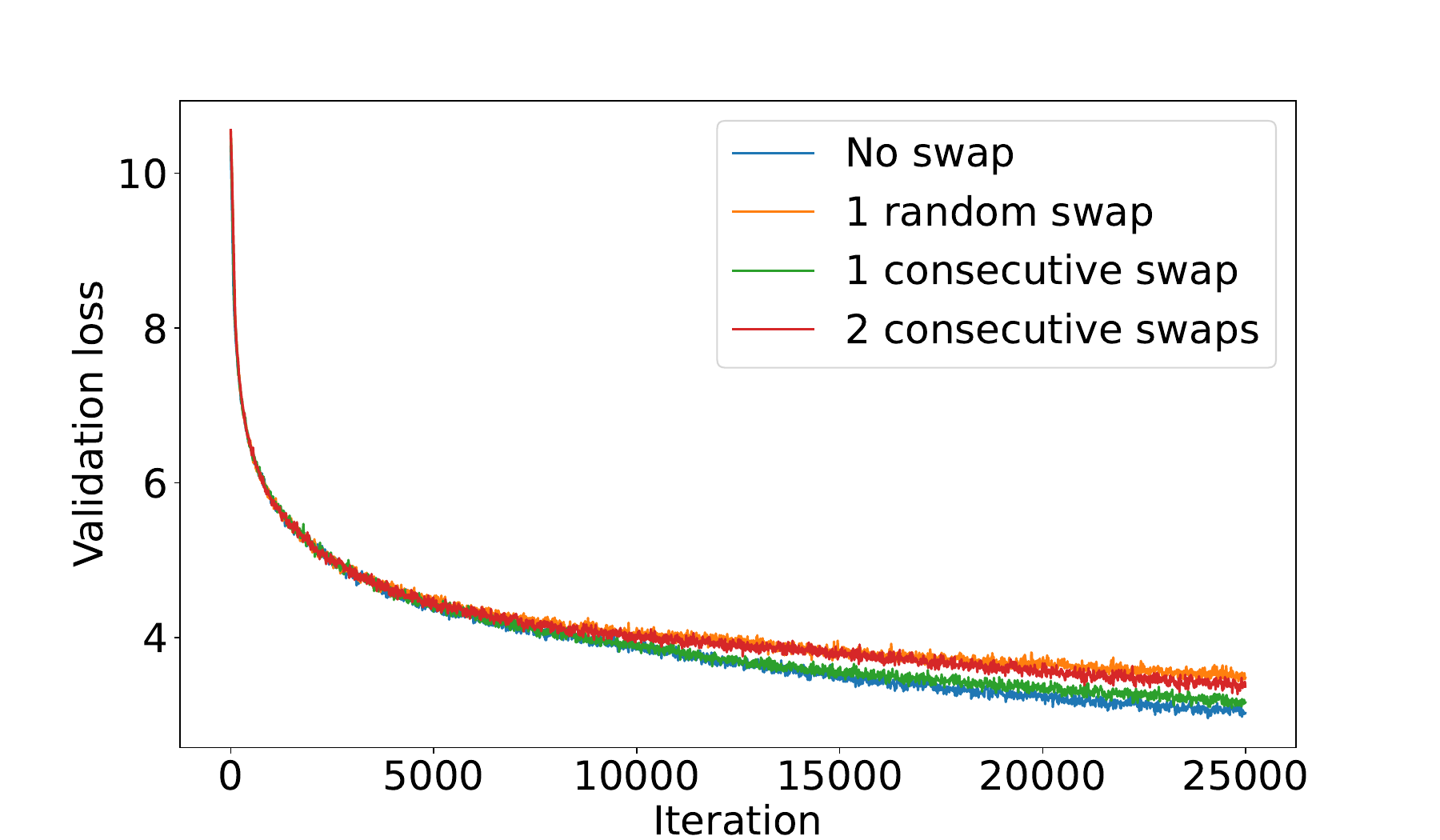} \\
(a) Impact of skipped layer selection. & (b) Impact of stage swapping on full model.\\	
\end{tabular}			
\caption{Convergence of LLaMa-30M model. The validation loss is calculated for the whole model for every 50th iteration.}
	\label{fig:motivation}
\end{figure*}

\textbf{Convergence Constraints}. To study the effects of stage skipping and swapping on the LLM convergence, we train a LLaMa-30M model (12 layers) divided in 6 stages with 2 layers each on the TinyStories dataset with 5 microbatches of size of 32 samples in two sets of experiments, summarized in Figure~\ref{fig:motivation}. 
 In Figure~\ref{fig:motivation} (a), we vary the selection of which stage to skip (for skipping percentage of 25\%): random, random with no skipping the first stage, and round robin with no skipping the first stage (skip each intermediate stage equal number of times). By comparing random and random with no skipping the first stage cases, we observe that the first stage is more critical than other stage and should not be skipped. Similar effect is also observed for larger transformer architectures \cite{vitrob,LayerDrop} and architectures with residual connections \cite{skipconnresnet}. Additionally, when we compare random and round-robin cases, we see that convergence is better when each intermediate stage is skipped uniformly and trained for an equal number of microbatches. Figure~\ref{fig:motivation} (b) shows that swapping execution order of two consecutive stages has negligible effect on the training loss, and swapping multiple stages or stages that are not consecutive causes more degradation.

Combining the aforementioned observations, we derive the following \textbf{C}onvergence \textbf{C}onstraints for our path selection:
\begin{itemize}
	\item \cc{1}: A path $p_i$ never skips the first stage, i.e., $S_{i_1} = S_0 \; \forall p_i \in \mathcal{P}$.\footnote{Our scheduler also works when multiple stages hold the critical layers - for simplicity we refer to them collectively as \(S_0\).}
	\item \cc{2}: A path $p_i$ may run out of order at most two consecutive stages (1 swap), i.e., for a path $p_i= S_{i_1},\ldots,S_{i_l}$,  $|{i_j} - {i_{j+1}}|  \leq 1 \;  \forall j \in (1,l)$.
	\item \cc{3}: Each stage $S_i$ ($i\geq1$) is skipped for an equal amount of paths.   
\end{itemize}

\textbf{Throughput Constraints.}
In a standard pipeline training, the whole model is executed sequentially and each node needs to receive activations of the microbatches from only one other node (the one before it) in the forward pass, and similarly for the backward pass as well. 
In other words, each node receives only one microbatch to process from each direction, unless one of the nodes is significantly slower than the consecutive one.\footnote{Note that it is possible to receive both forward and backward activations at the same time, and these can be handled in 1F1B manner~\cite{pipedream} where the backward pass is prioritized in the execution order.} 
However, as we introduce skips (and potentially swaps) in execution, it is possible for a node to simultaneously receive two microbatches from two different stages in the same direction. We refer to such cases as \textbf{collisions}, which can significantly degrade the end-to-end latency of a batch, which is the duration between the starting time of the first microbatch and the end time of the last microbatch (including the node computing time and communication time across stages). To avoid collisions, we employ swaps to run stages out of order for a microbatch, thus utilising a currently idle node to reduce instantaneous overutilisation of another. 

In addition, because of the caching of the activations that is needed for the backward pass, the number of active microbatches going through each node is limited by the memory of a node and denoted by (\(m\)). Overall, we impose two \textbf{T}hroughput \textbf{C}onstraints:
\begin{itemize}
    \item \tc{1}: At most \(m\) paths can go through each node \(N_i\). 
    \item \tc{2}: Minimize collisions via skipping or swapping the pipelining order.
\end{itemize}

\textbf{Problem Formulation.} We formalise the optimization problem of partial pipeline scheduler as follows: For a given network of $\mathcal{N}$ nodes with bandwidth and latency matrices $(\mathcal{B},\Lambda)$ and a LLM model consisting of pipeline stages $\mathcal{S}$, the number of microbatches $M$ and limitation of active microbatches $m$ per iteration, the partial pipeline scheduler aims to find the paths $\mathcal{P}$ that minimizes the maximum end-to-end latency across all microbatches of a given iteration:
\begin{align*}
&\mathcal{P} \leftarrow \texttt{Scheduler}(\mathcal{N},\mathcal{B},\Lambda,\mathcal{S},M,m) \\  
&\text{such that} \; \mathcal{P} := \underset{\mathcal{P} \in \mathcal{P}_{ALL}, \; \forall p_i \in \mathcal{P} }{\arg \min \max E2E(p_i)} \\
& \text{with constraints \cc{1}, \cc{2}, \cc{3}, \tc{1} and \tc{2}}
\end{align*}
where $E2E(\cdot)$ is the end-to-end latency of a microbatch where the starting time of a microbatch is also taken into account, and $\mathcal{P}_{ALL}$ is the set of all possible sets of paths. Forming the paths is itself an NP-hard problem (as detailed in Section~\ref{sec:partial_pipeline}). We thus split the problem into two parts: first allocation nodes in stages under full pipeline schedule and then finding the partial pipeline schedule for microbatches under the given node-stage mapping.

\subsection{ Allocating Nodes to Stages}
For a given network of $\mathcal{N}$ nodes (with bandwidth and latency matrices $(\mathcal{B},\Lambda)$) and the pipeline stages $\mathcal{S}$, the initial node arrangement matches each node with a stage for standard full and sequential pipelining (no skips or swaps).

This problem is already analyzed for heterogeneous networks in DT-FM~\cite{dtfm}, solved through 
a two-phase optimiser: clustering of nodes for DP and then arrangement of the connections for PP.  
DP clustering can be seen as \textit{graph partition problem} where each cluster corresponds to a stage and the partition cost is bounded by the slowest communication between two nodes in the same stage.
This problem can be solved via genetic algorithms as described in~\cite{dtfm}.
Then these clusters are ordered for PP, which can be represented as
an \textit{open-loop Traveling Salesman problem} \cite{TSP}.

To allocate nodes to stages in \sys, we modify the algorithm given in~\cite{dtfm} for the  unbalanced cluster sizes.
Following convergence constraints \cc{1} and \cc{3}, the initial stage is never skipped whereas all other stages are skipped equally, so that $k\%$ of the stages are skipped for each microbatch.

Assume the nodes allocated in a stage, as $S_i(\mathbf{n})$, we formulate the number of nodes per stages regarding the following equation:

\begin{align}
|S_i(\mathbf{n})| = |S_0(\mathbf{n})| \left( 1- \frac{s}{s-1}\cdot \frac{k}{100} \right)  \; \forall i \in (1,s).
\label{eqn:node_distribution}
\end{align}

To balance the workload across stages, we allocate the nodes per stage using the ratio given above.  
Thus, the procedure employed here is the same with DTFM~\cite{dtfm}, with the differences that one stage is larger than the others and we use a closed-loop TSP, as we require that the loss is computed again on Stage 0. With the optimised arrangement of nodes in stages, we can look for paths through the system that would satisfy our constraints.

\subsection{Partial and Reordered Pipelining}\label{sec:partial_pipeline}
\par Once nodes are arranged into stages, we schedule the microbatches through the system by skipping and reordering stages, which is the core of \sys. It is important to note the difference between a path and a microbatch. While a microbatch does travel down a path, multiple microbatches may use the same path. For example, when a node completes a backwards pass for a given microbatch, it can reuse the path it had just traversed, as it is the one that immediately has nodes with free memory. Thus we find a set of paths for the first wave of microbatches and reuse them a number of times during an iteration to meet the desired batch size.

 Given our problem formulation, we model the problem as a \textit{continuous-time Multi Agent Path Finding} (MAPF) \cite{ccbs} problem. In continuous-time MAPF, we are given a graph and a number of agents and each agent has a starting location and a desired end location. In our setting, we map the graph to the node connections after node allocation where the edges between nodes reflect the communication cost regarding the bandwidth and latency values of the corresponding nodes. Each agent represents a microbatch which travel from a starting node in stage \(S_0\) to the same destination node while passing $s(100-k)/100$ nodes in total.
 An agent can either wait at a node, move through the node (computation), or move to a different node, via some edge connecting the two. Each move is associated with a given (communication) cost. In the continuous-time setting, actions do not take 1 unit of time, but can be of arbitrary length. The problem has the additional constraints that no two agents can collide (be on the same node at the same time or the same edge). Since we assume full-duplex links, we do not concern ourselves with collisions on edges and agents can perform the wait action at the end of a link (or a node's buffer). But, since nodes have real physical limitations, we allow traversal of only one agent at a time through a node (constraint \tc{2}). 
 To find a viable solution we employ a modified version of the \textit{continuous-time Conflict-Based Search} (CBS)~\cite{ccbs} based on the changes described above.

 The first four constraints (\cc{1}, \cc{2}, \cc{3}, \tc{1}) are merely about finding the paths, while constraints \tc{2} deal with conflicts between two agents. 
 \cc{1} and \cc{2} constraints are individual per agent and thus can be managed by an A* search \cite{astar}, an exhaustive graph traversal algorithm. We use A* instead of the Safe interval path planning used in \cite{ccbs}, so that we can model the skips, swaps, and the additional constraints better.
However constraints \cc{3} and \tc{1} require inter-agent optimization 
because they specify global constraints - no more than this number of agents can go through this node in an iteration. This requires knowing all other agent's paths and thus existing solvers are insufficient. We thus delegate all constraints, apart from \cc{1} and \cc{2} to be resolved by CBS (with for \cc{3} and \tc{1} setting a constraint that an agent cannot visit all nodes in a stage or a specific node respectively, from \((-\inf,\inf)\). However, this proves extremely costly for large graphs or large number of agents, as an exponential number of possible solutions would need to be explored, before resolving \tc{2} constraints.

 We thus approximate the optimal solution, by employing several greedy heuristics, denoted by \hr{}. \hr{1}: When finding solutions we first employ CBS to find a number of solutions that satisfy \cc{1}, \cc{2} and \cc{3} constraints (32 in our experiments, as this proved sufficient to find good solutions, without expanding the subsequent search space too much). \hr{2}: For \cc{3} constraints we found it best to exclude from adding constraints for the \(\frac{|\mathcal{P}|}{4}\) slowest agents, as these would be the fastest paths they can take and any change would detrimentally affect the slowest path. Then for these generated solutions we solve for \tc{1} constraints. Here we again exclude the slowest path through a node from adding constraints for it. Once no \tc{1} constraints are detected, \tc{2} constraints are checked. Since we only concern ourselves with the critical path (the one that takes the longest, $i := {\arg \max E2E(p_i)} \; \forall p_i \in \mathcal{P}$), we priorities conflicts that occur with it. \hr{3}: For conflicts with paths faster than the \(\frac{|\mathcal{P}|}{4}\)th path, we only add constraints for the faster path. A constraint \tc{2} is added for each relevant agent by specifying that they cannot visit the conflicting node for the duration the other agent is traversing it.

\subsection{Path finding} 

The pseudocode of our path selection method is given in Algorithm~\ref{alg:csf_pseudocode}, the detailed steps can be found in Appx.~\ref{sec:appx_algorithm} in Algorithm~\ref{alg:csf}.
To find a set of paths satisfying the current constraints, as per CBS \cite{cbs}, we employ A* for each agent with a time dimension. When an agent travels between two nodes, its time is increased by the time it takes for a microbatch to travel down that link. Whenever an agent travels through a node, its time is increased by the time it takes to process a microbatch. If an agent is to visit a node and during the processing time there is a conflict that prohibits the agent from being in that node, its time of visiting the node is delayed to the end of conflict. 
\par An agent must skip exactly \(k\%\) of the stages. Thus when expanding a node, we do not consider the starting node until this condition is met. When we visit again the starting node the time of forward and backward, given all conflicts, for the given time is estimated and the node is readded to the heap with that cost and a special flag marking it as a potential final solution. When a node marked as a potential solution is popped from the heap, it is returned as the current fastest path for that agent that satisfies all current constraints.
\begin{algorithm}
\caption{Pseudocode of Path Selection Function.}
\label{alg:csf_pseudocode}
\begin{algorithmic}[1]
{\small
\REQUIRE{ \(\mathcal{S}\), \(k\%\), $G$ - initial node/stage arrangement}
\ENSURE{\(\mathcal{P}\)}
\STATE \(\mathcal{O} \leftarrow \emptyset\) 
\STATE \(Open\leftarrow\) Path assignment with no constraints
\WHILE{ \(|\mathcal{O}| < 32\)}
\STATE \(T \leftarrow \) best solution from \(Open\)
\IF{Violation of constraint \cc{3} in $T$}
\STATE Add constraints for fastest paths 
\STATE Find paths with new constraint via A*(\(G\))
\STATE Re-add to \(Open\)
\ELSE
\STATE \(\mathcal{O} \leftarrow \mathcal{O} \cup T\)
\ENDIF
\ENDWHILE

\WHILE{\(\mathcal{O}\) not empty}
\STATE \(T \leftarrow \) best solution from \(\mathcal{O}\)
\IF{Violation of constraint \tc{1} or \tc{2} in \(T\)}
\STATE Add constraints for violating fastest paths to \(T\)
\STATE Find paths with new constraint for \(T\) via A*(\(G\))
\STATE Re-add to \(\mathcal{O}\)
\ELSE
\STATE \textbf{Return} $\mathcal{P} \leftarrow T$
\ENDIF
\ENDWHILE

\STATE \textbf{Return} $\emptyset$
}
\end{algorithmic}
\end{algorithm}

\par Unlike traditional A* we do not make use of a visited set - we may consider a node during our search multiple times. This is because how we reach the starting node in the forward pass (which is what A* finds essentially), may not be the fastest way to do a forward + backward pass (which is why we read the starting node with the special flag). When expanding an A* node, we exclude all nodes that have been on that path or belong to a stage that has been visited from the set of potential next nodes. We may perform at most 1 swap in the ordering of stages for a given path (\cc{2} constraint).  Nodes that would go over the limit set by \cc{2} are excluded from consideration.

%% file: evaluation.tex
\section{Experimental Results of \sys}

We demonstrate that \sys provides significant improvement in the training throughput, while preserving convergence even when a model is trained partially. We evaluate several LLaMa-like models, ranging from 500M to 8B in a geo-distributed setting and we use RedPajamas~\cite{redpajama} and TOPv2 datasets~\cite{topv2}. We observe that using \sys, the models converge at the same rate but with a significantly higher throughput, meaning that training converges much faster in wall-clock time.

We present our experiments in two categories: throughput and convergence analysis.
Throughput experiments investigate the speed up of our partial pipeline scheduler \sys wrt. the baseline SOTA schedulers on a LLaMa-1.5B model. 
In convergence experiments, we analyse the convergence of training from scratch of LLaMa-500M model and parameter efficient finetuning of LLaMa-8B model with three skipping ratios: 0\%, 25\% and 33\%.

Code is available at \url{https://github.com/gensyn-ai/skippipe}, which utilises DecCom,\footnote{\url{https://github.com/NikolayBlagoev/DecCom-Python}} a modular stack framework for decentralised communication.

\subsection{Throughput}
We evaluate throughput improvement of our algorithm by measuring the end to end time pipeline training of an iteration.
We test the throughput of a LLaMa-1.5B model training (see Appendix \ref{appendix:mc} for architecture details) with 3 different skipping ratios (0\%, 25\% and 33\%) and different number of nodes. 
We utilise H100s to simulate the nodes for our measurement where we host several homogeneous nodes within the single GPU.
For simulating geo-distributed nodes, we utilise the bandwidth and latency values given in DT-FM~\cite{dtfm}.

\begin{figure}[tp!]
 \centering
    \begin{subfigure}{0.45\textwidth}
        \includegraphics[width=\textwidth]{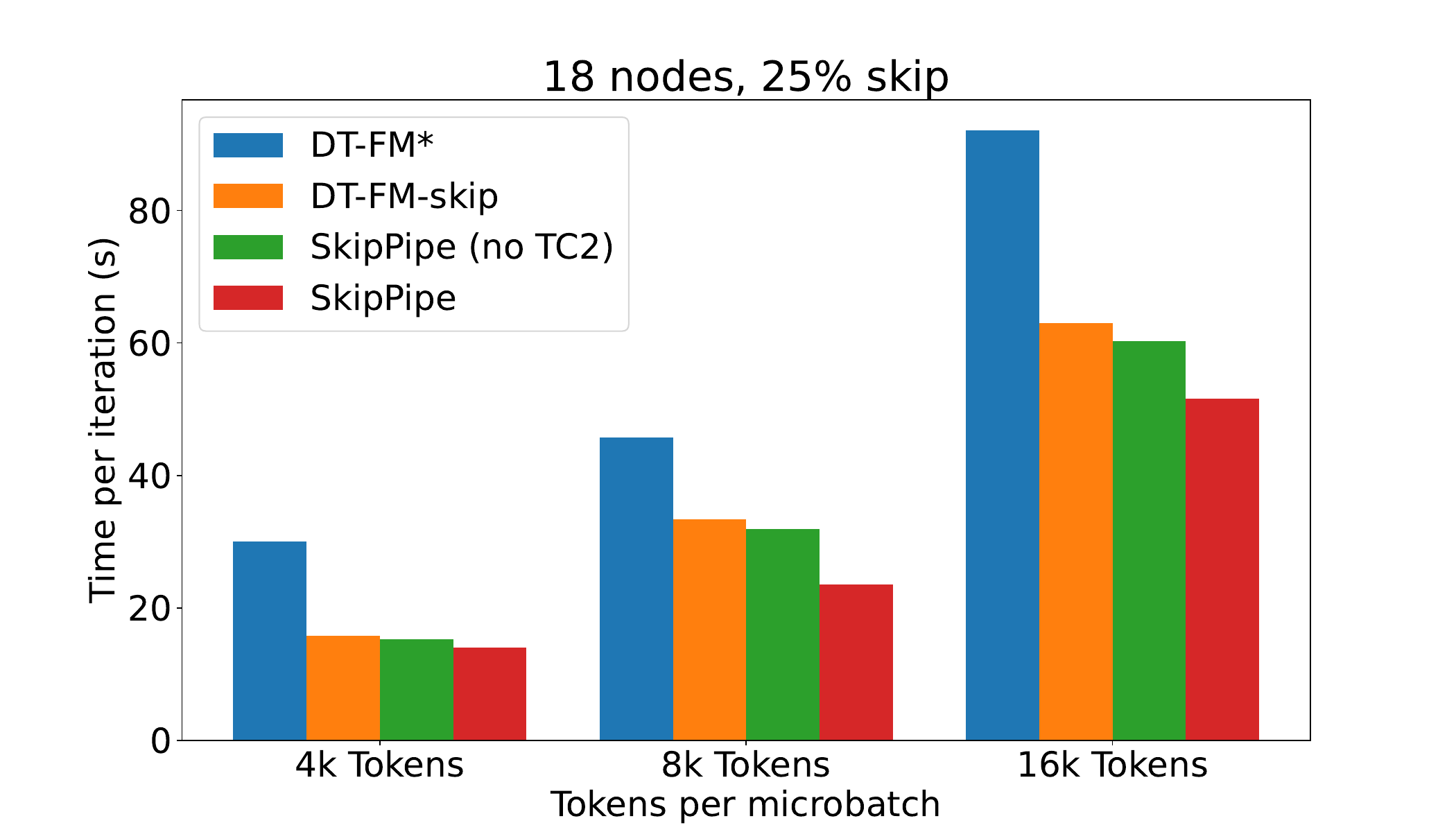}
        \caption{Heterogeneous setting with 18 nodes and 25\% skip rate.}
        \label{fig:throughput_first}
    \end{subfigure}
    \hfill
    \begin{subfigure}{0.45\textwidth}
        \includegraphics[width=\textwidth]{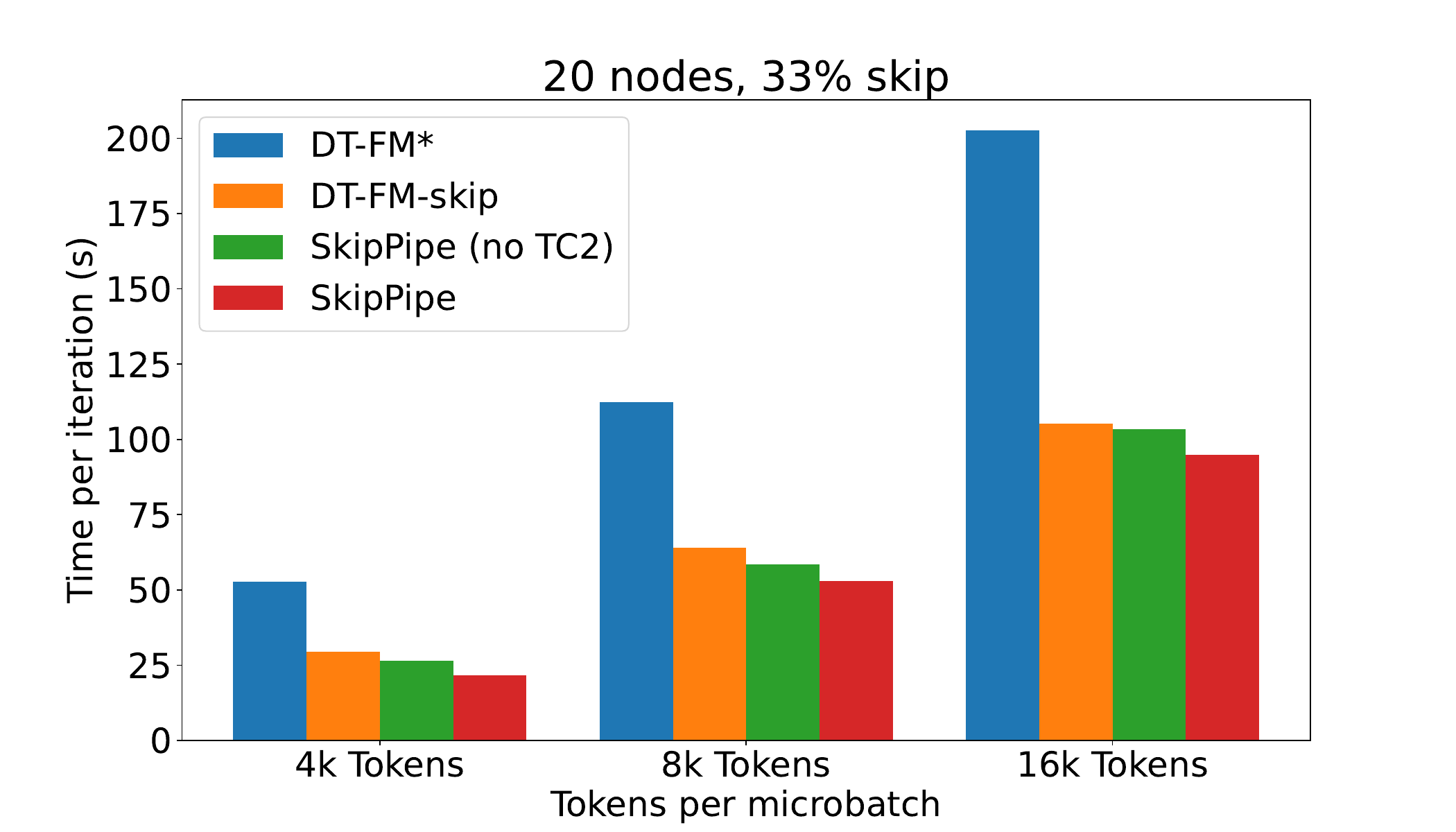}
        \caption{Heterogeneous setting with 20 nodes and 33\% skip rate.}
        \label{fig:throughput_second}
    \end{subfigure}
    \hfill
    \caption{Time per iteration with different strategies. We analyse four schedulers with two skip percentages (25\% and 33\%) and three token numbers (4K, 8K and 16K). \sys is compared with: DT-FM$^*$ representing the compensated results for the baseline with no skips, DT-FM-skip uses node arrangement of DT-FM and skips $k\%$ with additional constraints (see Appendix \ref{qrps}), \sys (no \tc{2}) is our scheduler without \tc{2}. }\label{fig:throughput}
\end{figure}

In Figure~\ref{fig:throughput}, we present the experimental results for two skip percentages ($k:=$25\% and 33\%) and 4 different schedulers. 
Also, we use varying number of samples per microbatch - of 1, 2 and 4, and make use of gradient accumulation.
We compare our scheduler, \sys, with (i) DT-FM: 0\% skip training using DT-FM scheduler, (ii) DT-FM-skip: $k$\% skip training using DT-FM scheduler with additional constraints (see Appendix \ref{qrps}), (iii) \sys (no \tc{2}): $k$\% skip training using our scheduler \sys where the collision constraint \tc{2} is ignored.
The time per iteration values are averaged over 50 different (randomly sampled) bandwidth and latency values.
Since we optimise the pipelining schedule for a given node/stage allocation, we measure the pipelining time and omit the data parallelism part where weight aggregation happens because the aggregation time is the same for a fixed node/stage allocation regardless of the microbatch paths. 
Finally, we perform one warm-up iteration where nodes discover each other. 

In Figure~\ref{fig:throughput_first}, we have the results for 25\% skip case. We tested 4 stages with 18 nodes where the nodes are distributed to the stages according to Equation~\ref{eqn:node_distribution}: $(6,4,4,4)$, except the 0\% skipping case used in DT-FM baseline.
To keep the node/stage sizes the same, for the DT-FM baseline, we use 16 nodes where nodes are equally distributed $(4,4,4,4)$.
To (over)compensate the baseline case using less nodes, we project their performance by proportionally reducing the end-to-end latency. 
Specifically, we multiply the latency of baseline by $\frac{16}{18}$, and these compensated latency results are represented by DT-FM$^*$.
Note that considering the communication of those additional nodes being ignored, this is an upper bound of their performance. 
As seen~\ref{fig:throughput_first}, \sys achieves \(40-50\%\) improvement compared to the baseline in the 8K and 16K tokens case.

In Figure~\ref{fig:throughput_second}, we have the results for 33\% skip case where we tested 6 stages with 20 nodes.
Similarly to the above case, number of nodes per stage is $(5,3,3,3,3,3)$, except the baseline, which is compensated by multiplying the corresponding latency values with $\frac{18}{20}$. 
We observe a consistent speedup of \(50\%\) compared to DT-FM$^*$, and even a \(55\%\) speed up in the \(16K\) tokens per microbatch. Additionally, removal of collisions provides a speedup of \(10\%\).

\subsection{Convergence}
\label{sec:convg}
With convergence experiments, we show that our scheduler \sys does not degrade the convergence of the training compared to no skipping case. We verify this by training from scratch a LLaMa-500M on the RedPajamas data \cite{redpajama} and finetuning LLaMa-8B model~\cite{llama8b} with LoRA~\cite{lora} on the TOPv2 dataset~\cite{topv2} with three different skip rates - 0\% (baseline), \(25\%\), and \(33\%\) skips.

\begin{figure}[tp!]
 \begin{subfigure}{0.235\textwidth}    \includegraphics[width=\textwidth]{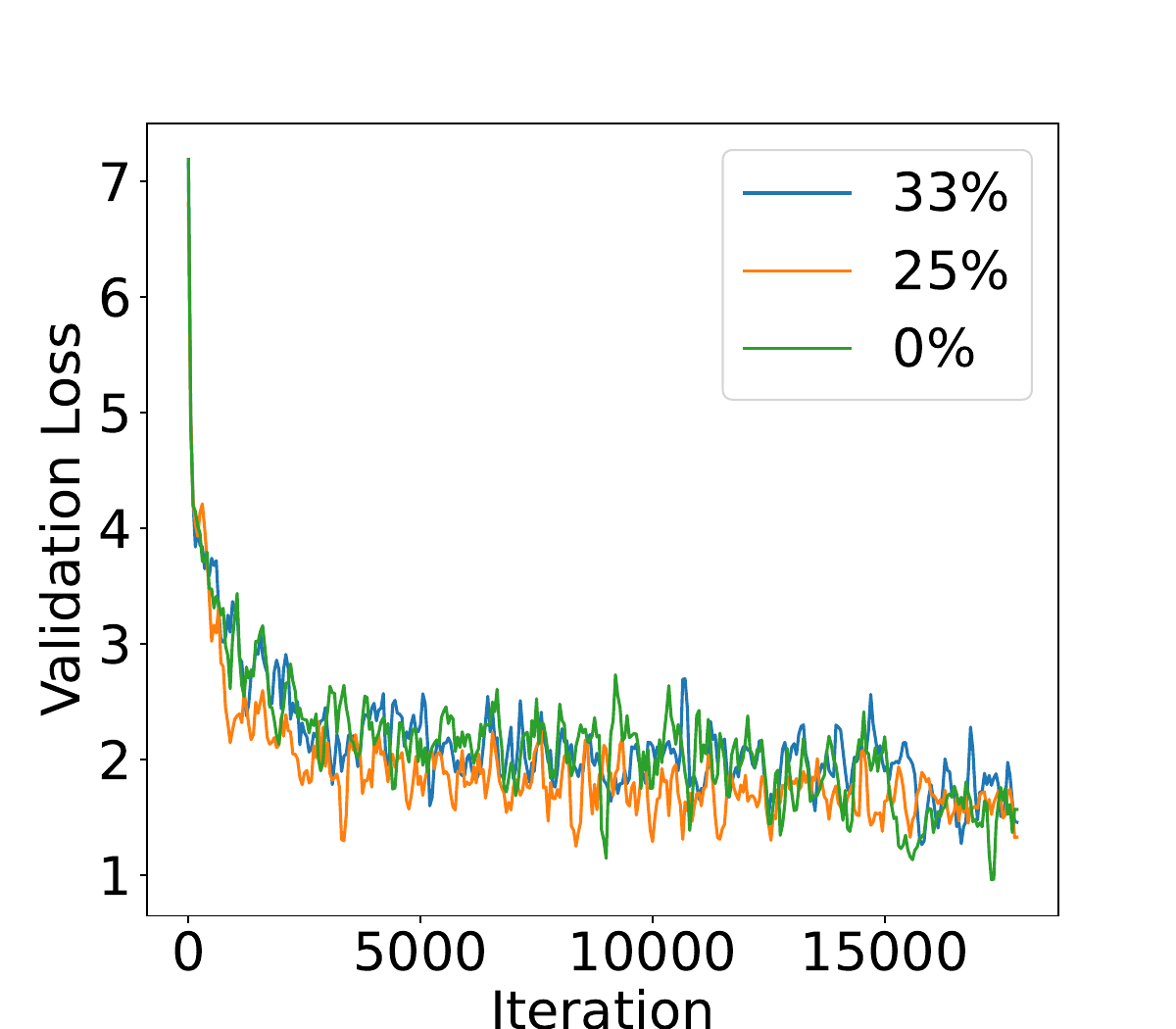}
        \caption{Training LLaMa-500M model with RedPajamas.}
        \label{fig:conv_first}
     \end{subfigure}
\hfill
 \begin{subfigure}{0.235\textwidth}
 \includegraphics[width=\textwidth]{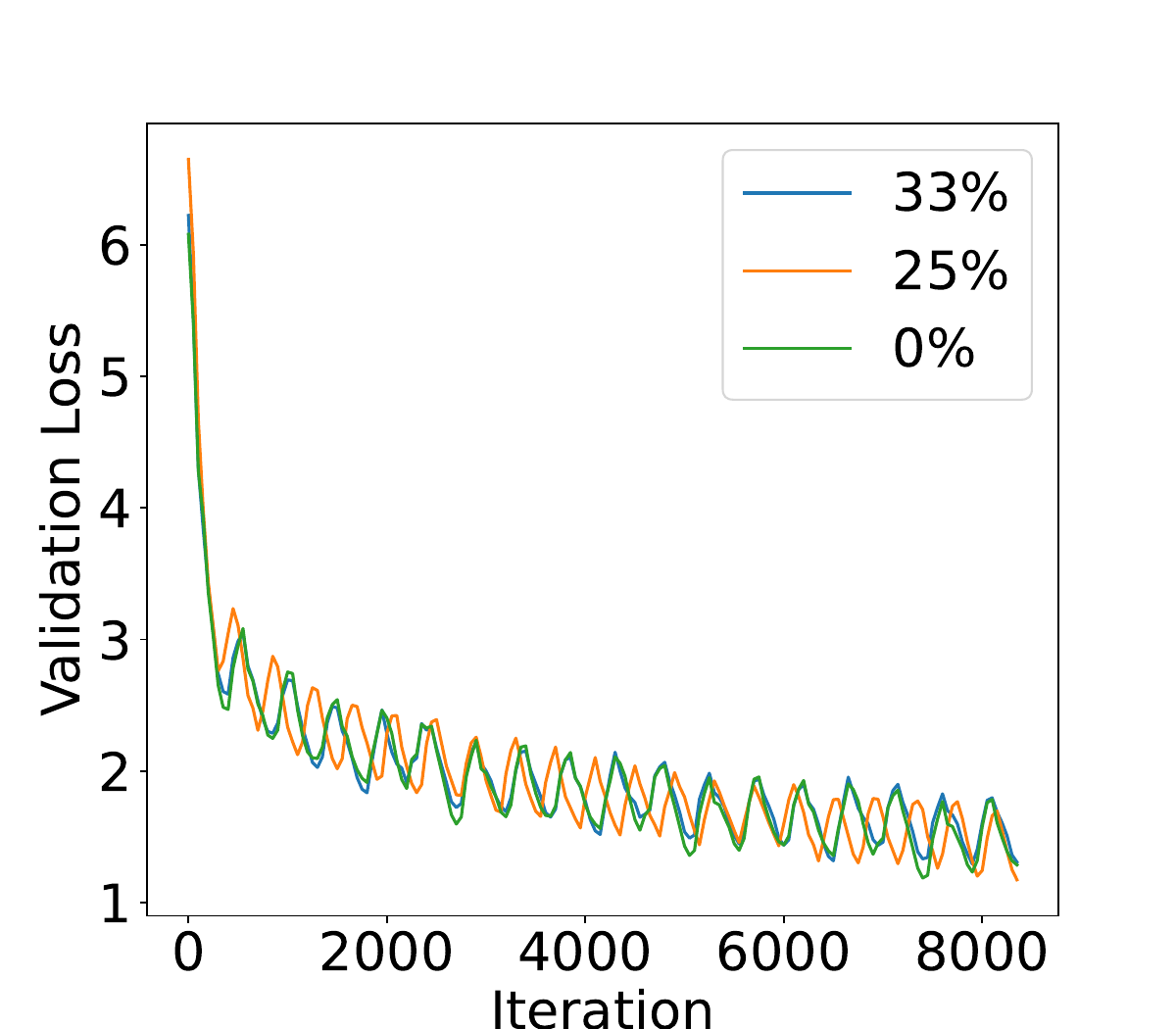}
        \caption{Finetuning LLaMa-8B model with TOPv2.}
        \label{fig:conv_second}
    \end{subfigure}
    \caption{Convergence of validation loss with \(33\%\) skip rate, \(25\%\) skip rate, and \(0\%\) skips (full model).}
    \label{fig:convg_ft} 
\end{figure}

In Figure~\ref{fig:convg_ft}, we report the validation loss every 50th iteration by running the entire model convergence (regardless of the training schedule). Our experiments show that \sys achieves similar convergence to the baseline for both training (see Figure~\ref{fig:conv_first}) and finetuning (see Figure~\ref{fig:conv_second}), despite training with a fraction of the model each time. Also, since \sys has a much higher throughput, convergence in terms of wall-clock time is significantly faster.

\section{Inference Benefits of \sys Training}
\par Training with partial pipelines results in models with inference robustness - they are resistant to a certain degree of layer/stage removal in inference, without sacrificing performance. We demonstrate this property in two settings: early exit where we employ self-speculative decoding to perform inference on the middle layer and fault tolerant inference where we test the inference pipeline with failing nodes.

\subsection{Early Exit}

\begin{table}[b]
\caption{Results for LLaMa-500M with various inference strategies. We report the ratio of accepted tokens at the middle layer and the relative speed up achieved with early exit.}
\label{table:resultsoptimal}
\begin{center}
\begin{threeparttable}
\begin{tabular}[H]{p{2.0cm} p{2.6cm} p{2.0cm}}
  \toprule
   & Token Acceptance  & Speedup    \\ 
  \midrule
   LayerSkip\tnote{a} & \(77.4\%\) & \textbf{1.76} \\ 
  \sys & \(62.8\%\) & 1.41 \\ 
  \bottomrule
\end{tabular}
{\small
\begin{tablenotes}
\item [a] Reported results in~\cite{LayerSkip} for LLaMa-1.5B.
\end{tablenotes}
}
\end{threeparttable}
\end{center}
\end{table}

\begin{table*}[ht]
  \caption{Perplexity (lower is better) on Arxiv dataset across 1000 evaluation samples for various inference and training skip rates. The inference (training) skip rate shows the percentage of stages being skipped during inference (training).}
  \label{table:ape}
  \begin{center}
  \begin{threeparttable}
    \begin{tabular}[H]{p{3cm}  p{0.7cm} p{0.7cm} p{0.7cm} p{0.7cm} p{0.7cm} p{0.7cm} p{0.7cm} p{0.7cm} p{0.7cm} p{0.7cm} p{0.7cm} p{0.7cm}}
      \toprule
   Inference skip rate   & \multicolumn{3}{c}{0\%} & \multicolumn{3}{c}{25\%} & \multicolumn{3}{c}{33\%}   & \multicolumn{3}{c}{50\%}\\
    Training skip rate & 0\% & 25\% & 33\%    & 0\% & 25\% & 33\% & 0\% & 25\% & 33\% & 0\% & 25\% & 33\% \\  
    \midrule
\textbf{Arxiv} $\downarrow$ &    8.59 & \textbf{8.32} & 8.96 &    29.58 & \textbf{8.5} & 9.8\tnote{a} &     33.35 & 10.57\tnote{a} & \textbf{9.0} &     81 & 9.9  & \textbf{9.57}\\

\bottomrule
    \end{tabular}
    \begin{tablenotes}
\item [a] Partial stage skips where number of stages is not divisible by the desired skip ratio.
\end{tablenotes}
\end{threeparttable}
  \end{center}
\end{table*}

\par Using our LLaMa-500M model trained with \(33\%\) skip rate, we employ self-speculative decoding strategies as in~\cite{LayerSkip}. To this end, during inference, we generate a number of draft tokens \(T_1, T_2, ...\) by stopping at the middle stage. These are then verified in a single pass by the remainder of the model with the middle stage's logits of the last draft token fed into the remaining stages. All tokens up to the first one that doesn't match between the draft tokens and the verified ones are kept. The first mismatched token is added to the prompt and generation continues from there. We compare the performance of our strategy against LayerSkip~\cite{LayerSkip}. Results are presented in Table~\ref{table:resultsoptimal}. 
We achieve 1.41 speedup for LLaMa-500M model without manipulating the loss function, whereas LayerSkip achieves 1.76 speedup for LLaMa-1.5B model by applying early exit loss function.

\subsection{Fault Tolerant Inference}

By training with \sys, the models exhibit robust inference results even if some stages are failed (except the first one). We demonstrate this by evaluating the perplexity of the trained Llama-500M models (in Section \ref{sec:convg}) given different inference stage skip rates on the Arxiv \cite{arxiv} dataset. For each skip rate a corresponding number of stages is dropped at random per sample. For the cases where the number of stages is not divisible by the desired skip ratio, we allow partial stage skips (executing a subset of the layers on the stage - the first half). 

As seen in Table \ref{table:ape}, our partial pipelining provides robustness against arbitrary stage removal during inference time. 
Overall, we observe relatively low perplexity values for the chosen dataset, as the models primarily trained on the Arxiv subset of the RedPajamas dataset.
Nonetheless, perplexities of the models trained with \sys stays lower than 10, whereas the baseline increases to 81 when half model is executed. 
Interestingly, we observe that when we perform partial stage skips, performance degrades more significantly. This suggests that layers exhibit a degree of co-learning. Thus it is possible to provide even stronger robustness by allowing for stages to be executed partially during training.

%% file: related.tex
\section{Related work}
\textbf{Efficient and Heterogeneity-aware Distributed Training.} 
There have been several works to improve (communication) efficiency of LLM training~\cite{diloco,demo,opendiloco}
where they show that the communication overhead can be significantly reduced by minimizing the synchronization for gradients in DP.
Moreover, there are several heterogeneity-aware scheduling methods~\cite{dtfm,swarm,hetpipe,metis,flashflex} proposing efficient DP and PP arrangement of the nodes to minimize the communication overhead. 
Yet, pipelining is always done in a sequential execution of all layers~\cite{gpipe,zb,pipedream,hetpipe}.
To the best of our knowledge, no prior work has studied the opportunity of optimizing for partial pipeline usage.

\textbf{Skip Connections and Early Exit.} Models employing skip connections have been known to exhibit robustness to random layer omission and perturbation \cite{skipconnresnet,vitrob}. Works such as \cite{stochasticdepth} demonstrated how larger models can be trained with less resources, by skipping certain layers during training. LayerDrop~\cite{LayerDrop} demonstrated that models trained partially are more robust to layer omission during inference. Based on this work, Layerskip~\cite{LayerSkip} proposed a novel training approach and loss function, which enabled them to perform early exiting during inference - running only part of the model to generate tokens and using the whole model only to verify their probability.

%% file: scheduling_paper.bbl
\begin{thebibliography}{34}
\providecommand{\natexlab}[1]{#1}
\providecommand{\url}[1]{\texttt{#1}}
\expandafter\ifx\csname urlstyle\endcsname\relax
  \providecommand{\doi}[1]{doi: #1}\else
  \providecommand{\doi}{doi: \begingroup \urlstyle{rm}\Url}\fi

\bibitem[Andreychuk et~al.(2021)Andreychuk, Yakovlev, Boyarski, and
  Stern]{ccbs}
Andreychuk, A., Yakovlev, K.~S., Boyarski, E., and Stern, R.
\newblock Improving continuous-time conflict based search.
\newblock In \emph{Thirty-Fifth {AAAI} Conference on Artificial Intelligence,
  {AAAI} 2021, Thirty-Third Conference on Innovative Applications of Artificial
  Intelligence, {IAAI} 2021, The Eleventh Symposium on Educational Advances in
  Artificial Intelligence, {EAAI} 2021, Virtual Event, February 2-9, 2021},
  pp.\  11220--11227. {AAAI} Press, 2021.
\newblock \doi{10.1609/AAAI.V35I13.17338}.
\newblock URL \url{https://doi.org/10.1609/aaai.v35i13.17338}.

\bibitem[Bhojanapalli et~al.(2021)Bhojanapalli, Chakrabarti, Glasner, Li,
  Unterthiner, and Veit]{vitrob}
Bhojanapalli, S., Chakrabarti, A., Glasner, D., Li, D., Unterthiner, T., and
  Veit, A.
\newblock Understanding robustness of transformers for image classification.
\newblock In \emph{2021 {IEEE/CVF} International Conference on Computer Vision,
  {ICCV} 2021, Montreal, QC, Canada, October 10-17, 2021}, pp.\  10211--10221.
  {IEEE}, 2021.
\newblock \doi{10.1109/ICCV48922.2021.01007}.
\newblock URL \url{https://doi.org/10.1109/ICCV48922.2021.01007}.

\bibitem[Brown et~al.(2020)Brown, Mann, Ryder, Subbiah, Kaplan, Dhariwal,
  Neelakantan, Shyam, Sastry, Askell, Agarwal, Herbert{-}Voss, Krueger,
  Henighan, Child, Ramesh, Ziegler, Wu, Winter, Hesse, Chen, Sigler, Litwin,
  Gray, Chess, Clark, Berner, McCandlish, Radford, Sutskever, and Amodei]{gpt3}
Brown, T.~B., Mann, B., Ryder, N., Subbiah, M., Kaplan, J., Dhariwal, P.,
  Neelakantan, A., Shyam, P., Sastry, G., Askell, A., Agarwal, S.,
  Herbert{-}Voss, A., Krueger, G., Henighan, T., Child, R., Ramesh, A.,
  Ziegler, D.~M., Wu, J., Winter, C., Hesse, C., Chen, M., Sigler, E., Litwin,
  M., Gray, S., Chess, B., Clark, J., Berner, C., McCandlish, S., Radford, A.,
  Sutskever, I., and Amodei, D.
\newblock Language models are few-shot learners.
\newblock In Larochelle, H., Ranzato, M., Hadsell, R., Balcan, M., and Lin, H.
  (eds.), \emph{NeurIPS}, 2020.
\newblock URL
  \url{https://proceedings.neurips.cc/paper/2020/hash/1457c0d6bfcb4967418bfb8ac142f64a-Abstract.html}.

\bibitem[Chen et~al.(2020)Chen, Ghoshal, Mehdad, Zettlemoyer, and Gupta]{topv2}
Chen, X., Ghoshal, A., Mehdad, Y., Zettlemoyer, L., and Gupta, S.
\newblock Low-resource domain adaptation for compositional task-oriented
  semantic parsing.
\newblock In Webber, B., Cohn, T., He, Y., and Liu, Y. (eds.),
  \emph{Proceedings of the 2020 Conference on Empirical Methods in Natural
  Language Processing, {EMNLP} 2020, Online, November 16-20, 2020}, pp.\
  5090--5100. Association for Computational Linguistics, 2020.
\newblock \doi{10.18653/V1/2020.EMNLP-MAIN.413}.
\newblock URL \url{https://doi.org/10.18653/v1/2020.emnlp-main.413}.

\bibitem[Chowdhery et~al.(2023)Chowdhery, Narang, Devlin, Bosma, Mishra,
  Roberts, Barham, Chung, Sutton, Gehrmann, Schuh, Shi, Tsvyashchenko, Maynez,
  Rao, Barnes, Tay, Shazeer, Prabhakaran, Reif, Du, Hutchinson, Pope, Bradbury,
  Austin, Isard, Gur{-}Ari, Yin, Duke, Levskaya, Ghemawat, Dev, Michalewski,
  Garcia, Misra, Robinson, Fedus, Zhou, Ippolito, Luan, Lim, Zoph, Spiridonov,
  Sepassi, Dohan, Agrawal, Omernick, Dai, Pillai, Pellat, Lewkowycz, Moreira,
  Child, Polozov, Lee, Zhou, Wang, Saeta, Diaz, Firat, Catasta, Wei,
  Meier{-}Hellstern, Eck, Dean, Petrov, and Fiedel]{palm}
Chowdhery, A., Narang, S., Devlin, J., Bosma, M., Mishra, G., Roberts, A.,
  Barham, P., Chung, H.~W., Sutton, C., Gehrmann, S., Schuh, P., Shi, K.,
  Tsvyashchenko, S., Maynez, J., Rao, A., Barnes, P., Tay, Y., Shazeer, N.,
  Prabhakaran, V., Reif, E., Du, N., Hutchinson, B., Pope, R., Bradbury, J.,
  Austin, J., Isard, M., Gur{-}Ari, G., Yin, P., Duke, T., Levskaya, A.,
  Ghemawat, S., Dev, S., Michalewski, H., Garcia, X., Misra, V., Robinson, K.,
  Fedus, L., Zhou, D., Ippolito, D., Luan, D., Lim, H., Zoph, B., Spiridonov,
  A., Sepassi, R., Dohan, D., Agrawal, S., Omernick, M., Dai, A.~M., Pillai,
  T.~S., Pellat, M., Lewkowycz, A., Moreira, E., Child, R., Polozov, O., Lee,
  K., Zhou, Z., Wang, X., Saeta, B., Diaz, M., Firat, O., Catasta, M., Wei, J.,
  Meier{-}Hellstern, K., Eck, D., Dean, J., Petrov, S., and Fiedel, N.
\newblock Palm: Scaling language modeling with pathways.
\newblock \emph{J. Mach. Learn. Res.}, 24:\penalty0 240:1--240:113, 2023.
\newblock URL \url{http://jmlr.org/papers/v24/22-1144.html}.

\bibitem[Clement et~al.(2019)Clement, Bierbaum, O'Keeffe, and Alemi]{arxiv}
Clement, C.~B., Bierbaum, M., O'Keeffe, K.~P., and Alemi, A.~A.
\newblock On the use of arxiv as a dataset, 2019.

\bibitem[Devlin et~al.(2019)Devlin, Chang, Lee, and Toutanova]{BERT}
Devlin, J., Chang, M., Lee, K., and Toutanova, K.
\newblock {BERT:} pre-training of deep bidirectional transformers for language
  understanding.
\newblock In Burstein, J., Doran, C., and Solorio, T. (eds.), \emph{Proceedings
  of the 2019 Conference of the North American Chapter of the Association for
  Computational Linguistics: Human Language Technologies, {NAACL-HLT} 2019,
  Minneapolis, MN, USA, June 2-7, 2019, Volume 1 (Long and Short Papers)}, pp.\
   4171--4186. Association for Computational Linguistics, 2019.
\newblock \doi{10.18653/V1/N19-1423}.
\newblock URL \url{https://doi.org/10.18653/v1/n19-1423}.

\bibitem[Doran \& Michie(1966)Doran and Michie]{astar}
Doran, J. and Michie, D.
\newblock {Experiments with the Graph Traverser Program}.
\newblock In \emph{Proc. of the Royal Society}, volume 294 Ser. A, 1966.

\bibitem[Douillard et~al.(2023)Douillard, Feng, Rusu, Chhaparia, Donchev,
  Kuncoro, Ranzato, Szlam, and Shen]{diloco}
Douillard, A., Feng, Q., Rusu, A.~A., Chhaparia, R., Donchev, Y., Kuncoro, A.,
  Ranzato, M., Szlam, A., and Shen, J.
\newblock Diloco: Distributed low-communication training of language models.
\newblock \emph{CoRR}, abs/2311.08105, 2023.

\bibitem[Dubey et~al.(2024)Dubey, Jauhri, Pandey, Kadian, Al{-}Dahle, Letman,
  Mathur, Schelten, Yang, Fan, Goyal, Hartshorn, Yang, Mitra, Sravankumar,
  Korenev, Hinsvark, Rao, Zhang, Rodriguez, Gregerson, Spataru, Rozi{\`{e}}re,
  Biron, Tang, Chern, Caucheteux, Nayak, Bi, Marra, McConnell, Keller, Touret,
  Wu, Wong, Ferrer, Nikolaidis, Allonsius, Song, Pintz, Livshits, Esiobu,
  Choudhary, Mahajan, Garcia{-}Olano, Perino, Hupkes, Lakomkin, AlBadawy,
  Lobanova, Dinan, Smith, Radenovic, Zhang, Synnaeve, Lee, Anderson, Nail,
  Mialon, Pang, Cucurell, Nguyen, Korevaar, Xu, Touvron, Zarov, Ibarra,
  Kloumann, Misra, Evtimov, Copet, Lee, Geffert, Vranes, Park, Mahadeokar,
  Shah, van~der Linde, Billock, Hong, Lee, Fu, Chi, Huang, Liu, Wang, Yu,
  Bitton, Spisak, Park, Rocca, Johnstun, Saxe, Jia, Alwala, Upasani, Plawiak,
  Li, Heafield, Stone, and et~al.]{llama8b}
Dubey, A., Jauhri, A., Pandey, A., Kadian, A., Al{-}Dahle, A., Letman, A.,
  Mathur, A., Schelten, A., Yang, A., Fan, A., Goyal, A., Hartshorn, A., Yang,
  A., Mitra, A., Sravankumar, A., Korenev, A., Hinsvark, A., Rao, A., Zhang,
  A., Rodriguez, A., Gregerson, A., Spataru, A., Rozi{\`{e}}re, B., Biron, B.,
  Tang, B., Chern, B., Caucheteux, C., Nayak, C., Bi, C., Marra, C., McConnell,
  C., Keller, C., Touret, C., Wu, C., Wong, C., Ferrer, C.~C., Nikolaidis, C.,
  Allonsius, D., Song, D., Pintz, D., Livshits, D., Esiobu, D., Choudhary, D.,
  Mahajan, D., Garcia{-}Olano, D., Perino, D., Hupkes, D., Lakomkin, E.,
  AlBadawy, E., Lobanova, E., Dinan, E., Smith, E.~M., Radenovic, F., Zhang,
  F., Synnaeve, G., Lee, G., Anderson, G.~L., Nail, G., Mialon, G., Pang, G.,
  Cucurell, G., Nguyen, H., Korevaar, H., Xu, H., Touvron, H., Zarov, I.,
  Ibarra, I.~A., Kloumann, I.~M., Misra, I., Evtimov, I., Copet, J., Lee, J.,
  Geffert, J., Vranes, J., Park, J., Mahadeokar, J., Shah, J., van~der Linde,
  J., Billock, J., Hong, J., Lee, J., Fu, J., Chi, J., Huang, J., Liu, J.,
  Wang, J., Yu, J., Bitton, J., Spisak, J., Park, J., Rocca, J., Johnstun, J.,
  Saxe, J., Jia, J., Alwala, K.~V., Upasani, K., Plawiak, K., Li, K., Heafield,
  K., Stone, K., and et~al.
\newblock The llama 3 herd of models.
\newblock \emph{CoRR}, abs/2407.21783, 2024.
\newblock \doi{10.48550/ARXIV.2407.21783}.
\newblock URL \url{https://doi.org/10.48550/arXiv.2407.21783}.

\bibitem[Elhoushi et~al.(2024)Elhoushi, Shrivastava, Liskovich, Hosmer, Wasti,
  Lai, Mahmoud, Acun, Agarwal, Roman, Aly, Chen, and Wu]{LayerSkip}
Elhoushi, M., Shrivastava, A., Liskovich, D., Hosmer, B., Wasti, B., Lai, L.,
  Mahmoud, A., Acun, B., Agarwal, S., Roman, A., Aly, A.~A., Chen, B., and Wu,
  C.
\newblock Layerskip: Enabling early exit inference and self-speculative
  decoding.
\newblock In Ku, L., Martins, A., and Srikumar, V. (eds.), \emph{Proceedings of
  the 62nd Annual Meeting of the Association for Computational Linguistics
  (Volume 1: Long Papers), {ACL} 2024, Bangkok, Thailand, August 11-16, 2024},
  pp.\  12622--12642. Association for Computational Linguistics, 2024.
\newblock \doi{10.18653/V1/2024.ACL-LONG.681}.
\newblock URL \url{https://doi.org/10.18653/v1/2024.acl-long.681}.

\bibitem[Fan et~al.(2020)Fan, Grave, and Joulin]{LayerDrop}
Fan, A., Grave, E., and Joulin, A.
\newblock Reducing transformer depth on demand with structured dropout.
\newblock In \emph{8th International Conference on Learning Representations,
  {ICLR} 2020, Addis Ababa, Ethiopia, April 26-30, 2020}. OpenReview.net, 2020.
\newblock URL \url{https://openreview.net/forum?id=SylO2yStDr}.

\bibitem[Foundation()]{wikidump}
Foundation, W.
\newblock Wikimedia downloads.
\newblock URL \url{https://dumps.wikimedia.org}.

\bibitem[Harlap et~al.(2018)Harlap, Narayanan, Phanishayee, Seshadri, Devanur,
  Ganger, and Gibbons]{pipedream}
Harlap, A., Narayanan, D., Phanishayee, A., Seshadri, V., Devanur, N.~R.,
  Ganger, G.~R., and Gibbons, P.~B.
\newblock Pipedream: Fast and efficient pipeline parallel {DNN} training.
\newblock \emph{CoRR}, abs/1806.03377, 2018.

\bibitem[Hu et~al.(2022)Hu, Shen, Wallis, Allen{-}Zhu, Li, Wang, Wang, and
  Chen]{lora}
Hu, E.~J., Shen, Y., Wallis, P., Allen{-}Zhu, Z., Li, Y., Wang, S., Wang, L.,
  and Chen, W.
\newblock Lora: Low-rank adaptation of large language models.
\newblock In \emph{The Tenth International Conference on Learning
  Representations, {ICLR} 2022, Virtual Event, April 25-29, 2022}.
  OpenReview.net, 2022.
\newblock URL \url{https://openreview.net/forum?id=nZeVKeeFYf9}.

\bibitem[Huang et~al.(2016)Huang, Sun, Liu, Sedra, and
  Weinberger]{stochasticdepth}
Huang, G., Sun, Y., Liu, Z., Sedra, D., and Weinberger, K.~Q.
\newblock Deep networks with stochastic depth.
\newblock In Leibe, B., Matas, J., Sebe, N., and Welling, M. (eds.),
  \emph{Computer Vision - {ECCV} 2016 - 14th European Conference, Amsterdam,
  The Netherlands, October 11-14, 2016, Proceedings, Part {IV}}, volume 9908 of
  \emph{Lecture Notes in Computer Science}, pp.\  646--661. Springer, 2016.
\newblock \doi{10.1007/978-3-319-46493-0\_39}.
\newblock URL \url{https://doi.org/10.1007/978-3-319-46493-0\_39}.

\bibitem[Huang et~al.(2019)Huang, Cheng, Bapna, Firat, Chen, Chen, Lee, Ngiam,
  Le, Wu, and Chen]{gpipe}
Huang, Y., Cheng, Y., Bapna, A., Firat, O., Chen, D., Chen, M.~X., Lee, H.,
  Ngiam, J., Le, Q.~V., Wu, Y., and Chen, Z.
\newblock Gpipe: Efficient training of giant neural networks using pipeline
  parallelism.
\newblock In \emph{NeurIPS}, pp.\  103--112, 2019.

\bibitem[Jaghouar et~al.(2024)Jaghouar, Ong, and Hagemann]{opendiloco}
Jaghouar, S., Ong, J.~M., and Hagemann, J.
\newblock Opendiloco: An open-source framework for globally distributed
  low-communication training.
\newblock \emph{CoRR}, abs/2407.07852, 2024.

\bibitem[Kudo \& Richardson(2018)Kudo and Richardson]{SP}
Kudo, T. and Richardson, J.
\newblock Sentencepiece: {A} simple and language independent subword tokenizer
  and detokenizer for neural text processing.
\newblock In Blanco, E. and Lu, W. (eds.), \emph{Proceedings of the 2018
  Conference on Empirical Methods in Natural Language Processing, {EMNLP} 2018:
  System Demonstrations, Brussels, Belgium, October 31 - November 4, 2018},
  pp.\  66--71. Association for Computational Linguistics, 2018.
\newblock \doi{10.18653/V1/D18-2012}.
\newblock URL \url{https://doi.org/10.18653/v1/d18-2012}.

\bibitem[Papadimitriou(1977)]{TSP}
Papadimitriou, C.~H.
\newblock The euclidean travelling salesman problem is np-complete.
\newblock \emph{Theoretical Computer Science}, 4\penalty0 (3):\penalty0
  237--244, 1977.
\newblock ISSN 0304-3975.
\newblock \doi{https://doi.org/10.1016/0304-3975(77)90012-3}.
\newblock URL
  \url{https://www.sciencedirect.com/science/article/pii/0304397577900123}.

\bibitem[Park et~al.(2020)Park, Yun, Yi, Nguyen, Lee, Choi, Noh, and
  Choi]{hetpipe}
Park, J.~H., Yun, G., Yi, C.~M., Nguyen, N.~T., Lee, S., Choi, J., Noh, S.~H.,
  and Choi, Y.
\newblock Hetpipe: Enabling large {DNN} training on (whimpy) heterogeneous
  {GPU} clusters through integration of pipelined model parallelism and data
  parallelism.
\newblock In Gavrilovska, A. and Zadok, E. (eds.), \emph{Proceedings of the
  2020 {USENIX} Annual Technical Conference, {USENIX} {ATC} 2020, July 15-17,
  2020}, pp.\  307--321. {USENIX} Association, 2020.
\newblock URL \url{https://www.usenix.org/conference/atc20/presentation/park}.

\bibitem[Peng et~al.(2024)Peng, Quesnelle, and Kingma]{demo}
Peng, B., Quesnelle, J., and Kingma, D.~P.
\newblock Decoupled momentum optimization.
\newblock \emph{arXiv preprint arXiv:2411.19870}, 2024.

\bibitem[Qi et~al.(2024)Qi, Wan, Huang, and Lin]{zb}
Qi, P., Wan, X., Huang, G., and Lin, M.
\newblock Zero bubble (almost) pipeline parallelism.
\newblock In \emph{The Twelfth International Conference on Learning
  Representations, {ICLR} 2024, Vienna, Austria, May 7-11, 2024}.
  OpenReview.net, 2024.
\newblock URL \url{https://openreview.net/forum?id=tuzTN0eIO5}.

\bibitem[Radford et~al.(2018)Radford, Narasimhan, Salimans, and Sutskever]{gpt}
Radford, A., Narasimhan, K., Salimans, T., and Sutskever, I.
\newblock Improving language understanding by generative pre-training, 2018.

\bibitem[Ryabinin et~al.(2023)Ryabinin, Dettmers, Diskin, and Borzunov]{swarm}
Ryabinin, M., Dettmers, T., Diskin, M., and Borzunov, A.
\newblock {SWARM} parallelism: Training large models can be surprisingly
  communication-efficient.
\newblock In Krause, A., Brunskill, E., Cho, K., Engelhardt, B., Sabato, S.,
  and Scarlett, J. (eds.), \emph{International Conference on Machine Learning,
  {ICML} 2023, 23-29 July 2023, Honolulu, Hawaii, {USA}}, volume 202 of
  \emph{Proceedings of Machine Learning Research}, pp.\  29416--29440. {PMLR},
  2023.
\newblock URL \url{https://proceedings.mlr.press/v202/ryabinin23a.html}.

\bibitem[Sharon et~al.(2015)Sharon, Stern, Felner, and Sturtevant]{cbs}
Sharon, G., Stern, R., Felner, A., and Sturtevant, N.~R.
\newblock Conflict-based search for optimal multi-agent pathfinding.
\newblock \emph{Artif. Intell.}, 219:\penalty0 40--66, 2015.
\newblock \doi{10.1016/J.ARTINT.2014.11.006}.
\newblock URL \url{https://doi.org/10.1016/j.artint.2014.11.006}.

\bibitem[Shoeybi et~al.(2019)Shoeybi, Patwary, Puri, LeGresley, Casper, and
  Catanzaro]{megatron}
Shoeybi, M., Patwary, M., Puri, R., LeGresley, P., Casper, J., and Catanzaro,
  B.
\newblock Megatron-lm: Training multi-billion parameter language models using
  model parallelism.
\newblock \emph{CoRR}, abs/1909.08053, 2019.
\newblock URL \url{http://arxiv.org/abs/1909.08053}.

\bibitem[Touvron et~al.(2023)Touvron, Lavril, Izacard, Martinet, Lachaux,
  Lacroix, Rozi{\`{e}}re, Goyal, Hambro, Azhar, Rodriguez, Joulin, Grave, and
  Lample]{llama}
Touvron, H., Lavril, T., Izacard, G., Martinet, X., Lachaux, M., Lacroix, T.,
  Rozi{\`{e}}re, B., Goyal, N., Hambro, E., Azhar, F., Rodriguez, A., Joulin,
  A., Grave, E., and Lample, G.
\newblock Llama: Open and efficient foundation language models.
\newblock \emph{CoRR}, abs/2302.13971, 2023.
\newblock \doi{10.48550/ARXIV.2302.13971}.
\newblock URL \url{https://doi.org/10.48550/arXiv.2302.13971}.

\bibitem[Um et~al.(2024)Um, Oh, Kang, Lee, Kim, Kim, Kim, Muzzammil, and
  Jeon]{metis}
Um, T., Oh, B., Kang, M., Lee, W., Kim, G., Kim, D., Kim, Y., Muzzammil, M.,
  and Jeon, M.
\newblock Metis: Fast automatic distributed training on heterogeneous gpus.
\newblock In \emph{{USENIX} {ATC}}, pp.\  563--578. {USENIX} Association, 2024.

\bibitem[Vaswani et~al.(2017)Vaswani, Shazeer, Parmar, Uszkoreit, Jones, Gomez,
  Kaiser, and Polosukhin]{transformer}
Vaswani, A., Shazeer, N., Parmar, N., Uszkoreit, J., Jones, L., Gomez, A.~N.,
  Kaiser, L., and Polosukhin, I.
\newblock Attention is all you need.
\newblock In Guyon, I., von Luxburg, U., Bengio, S., Wallach, H.~M., Fergus,
  R., Vishwanathan, S. V.~N., and Garnett, R. (eds.), \emph{Advances in Neural
  Information Processing Systems 30: Annual Conference on Neural Information
  Processing Systems 2017, December 4-9, 2017, Long Beach, CA, {USA}}, pp.\
  5998--6008, 2017.
\newblock URL
  \url{https://proceedings.neurips.cc/paper/2017/hash/3f5ee243547dee91fbd053c1c4a845aa-Abstract.html}.

\bibitem[Veit et~al.(2016)Veit, Wilber, and Belongie]{skipconnresnet}
Veit, A., Wilber, M.~J., and Belongie, S.~J.
\newblock Residual networks behave like ensembles of relatively shallow
  networks.
\newblock In Lee, D.~D., Sugiyama, M., von Luxburg, U., Guyon, I., and Garnett,
  R. (eds.), \emph{Advances in Neural Information Processing Systems 29: Annual
  Conference on Neural Information Processing Systems 2016, December 5-10,
  2016, Barcelona, Spain}, pp.\  550--558, 2016.
\newblock URL
  \url{https://proceedings.neurips.cc/paper/2016/hash/37bc2f75bf1bcfe8450a1a41c200364c-Abstract.html}.

\bibitem[Weber et~al.(2024)Weber, Fu, Anthony, Oren, Adams, Alexandrov, Lyu,
  Nguyen, Yao, Adams, Athiwaratkun, Chalamala, Chen, Ryabinin, Dao, Liang, Ré,
  Rish, and Zhang]{redpajama}
Weber, M., Fu, D., Anthony, Q., Oren, Y., Adams, S., Alexandrov, A., Lyu, X.,
  Nguyen, H., Yao, X., Adams, V., Athiwaratkun, B., Chalamala, R., Chen, K.,
  Ryabinin, M., Dao, T., Liang, P., Ré, C., Rish, I., and Zhang, C.
\newblock Redpajama: an open dataset for training large language models, 2024.
\newblock URL \url{https://arxiv.org/abs/2411.12372}.

\bibitem[Yan et~al.(2024)Yan, Jiang, Tao, Nie, Cui, and Yuan]{flashflex}
Yan, R., Jiang, Y., Tao, W., Nie, X., Cui, B., and Yuan, B.
\newblock Flashflex: Accommodating large language model training over
  heterogeneous environment.
\newblock \emph{CoRR}, abs/2409.01143, 2024.

\bibitem[Yuan et~al.(2022)Yuan, He, Davis, Zhang, Dao, Chen, Liang, R{\'{e}},
  and Zhang]{dtfm}
Yuan, B., He, Y., Davis, J., Zhang, T., Dao, T., Chen, B., Liang, P., R{\'{e}},
  C., and Zhang, C.
\newblock Decentralized training of foundation models in heterogeneous
  environments.
\newblock In Koyejo, S., Mohamed, S., Agarwal, A., Belgrave, D., Cho, K., and
  Oh, A. (eds.), \emph{Advances in Neural Information Processing Systems 35:
  Annual Conference on Neural Information Processing Systems 2022, NeurIPS
  2022, New Orleans, LA, USA, November 28 - December 9, 2022}, 2022.
\newblock URL
  \url{http://papers.nips.cc/paper\_files/paper/2022/hash/a37d615b61f999a5fa276adb14643476-Abstract-Conference.html}.

\end{thebibliography}
